%% file: main.tex
\documentclass[10pt,twocolumn,letterpaper]{article}

\usepackage{cvpr}              

\input{preamble}

\definecolor{cvprblue}{rgb}{0.21,0.49,0.74}
\usepackage[pagebackref,breaklinks,colorlinks,citecolor=cvprblue]{hyperref}

\usepackage{algorithmicx,algorithm}
\usepackage{bm}
\usepackage{multirow}
\usepackage{eqparbox}
\usepackage{color}
\usepackage{float}
\usepackage{makecell}
\usepackage[noend]{algpseudocode}
\usepackage{multirow}
\usepackage{graphbox}
\usepackage{array,booktabs}

\usepackage{comment}
\usepackage{stmaryrd}
\usepackage{breqn}

\makeatletter
\renewcommand{\ALG@beginalgorithmic}{\small} 
\makeatother

\title{3D-VirtFusion: Synthetic 3D Data Augmentation through Generative Diffusion Models and Controllable Editing}

\author{Shichao Dong$^{1,2}$ \quad Ze Yang$^{2}$ \quad Guosheng Lin$^{1,2}$ \thanks{Corresponding author: G.Lin (e-mail:gslin@ntu.edu.sg)}\\
	$^{1}$ S-lab, Nanyang Technological University, Singapore  \\ $^{2}$College of Computing and Data Science, Nanyang Technological University, Singapore  \\
	{\tt\small \{scdong, gslin\}@ntu.edu.sg} \quad {\tt\small \{ze001\}@e.ntu.edu.sg}
}

\begin{document}
\maketitle
\input{sec/0_abstract}    
\input{sec/1_intro}

\input{sec/2_relative}
\input{sec/3_method}

\input{sec/4_experiments}
\input{sec/5_conclusion}

\clearpage

{
    \small
    \bibliographystyle{ieeenat_fullname}
    \bibliography{main}
}

\end{document}


\maketitle

\begin{figure*}[t]
	\begin{center}
		\includegraphics[width=1.0\linewidth]{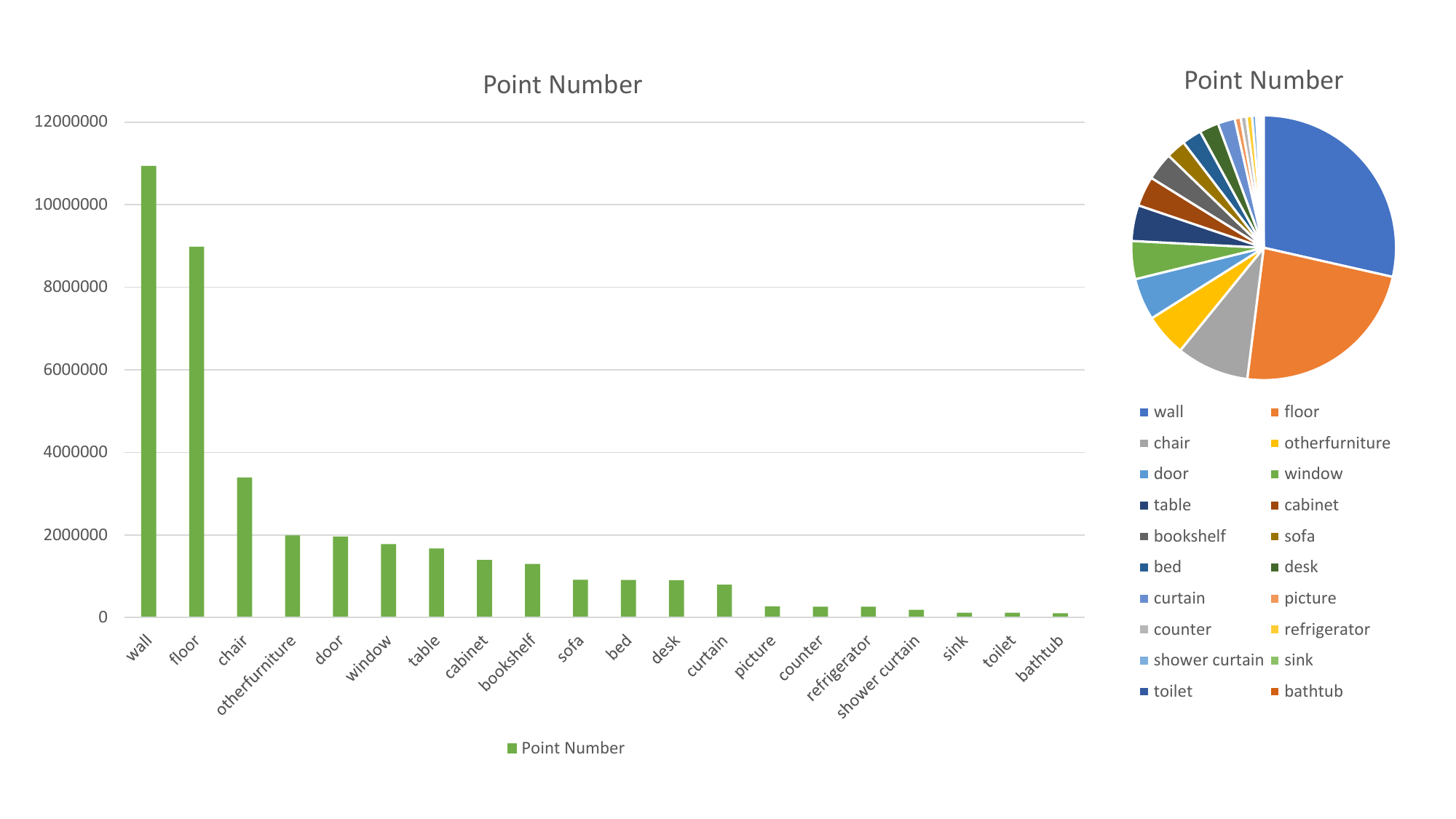}
	\end{center}
    \vspace{-8mm}
	\captionof{figure}{Statistics on point numbers of all semantic categories in ScanNet dataset.}
    \label{fig:Statistic}
\end{figure*}

\section{Data Distribution on ScanNet dataset}
 As shown in Figure \ref{fig:Statistic}, the data distribution within ScanNet exhibits a pronounced bias towards the predominant classes, such as wall, floor, and chair. Conversely, minority classes like sink, toilet, and bathtub comprise less than $1\%$ of the overall data points. Data imbalance can bias the model towards the majority classes, leading to inadequate learning and classification performance for minority classes. This can potentially lead to higher rates of false positives for majority classes and false negatives for minority classes.

\begin{figure*}[t]
	\begin{center}
		\includegraphics[width=0.9\linewidth]{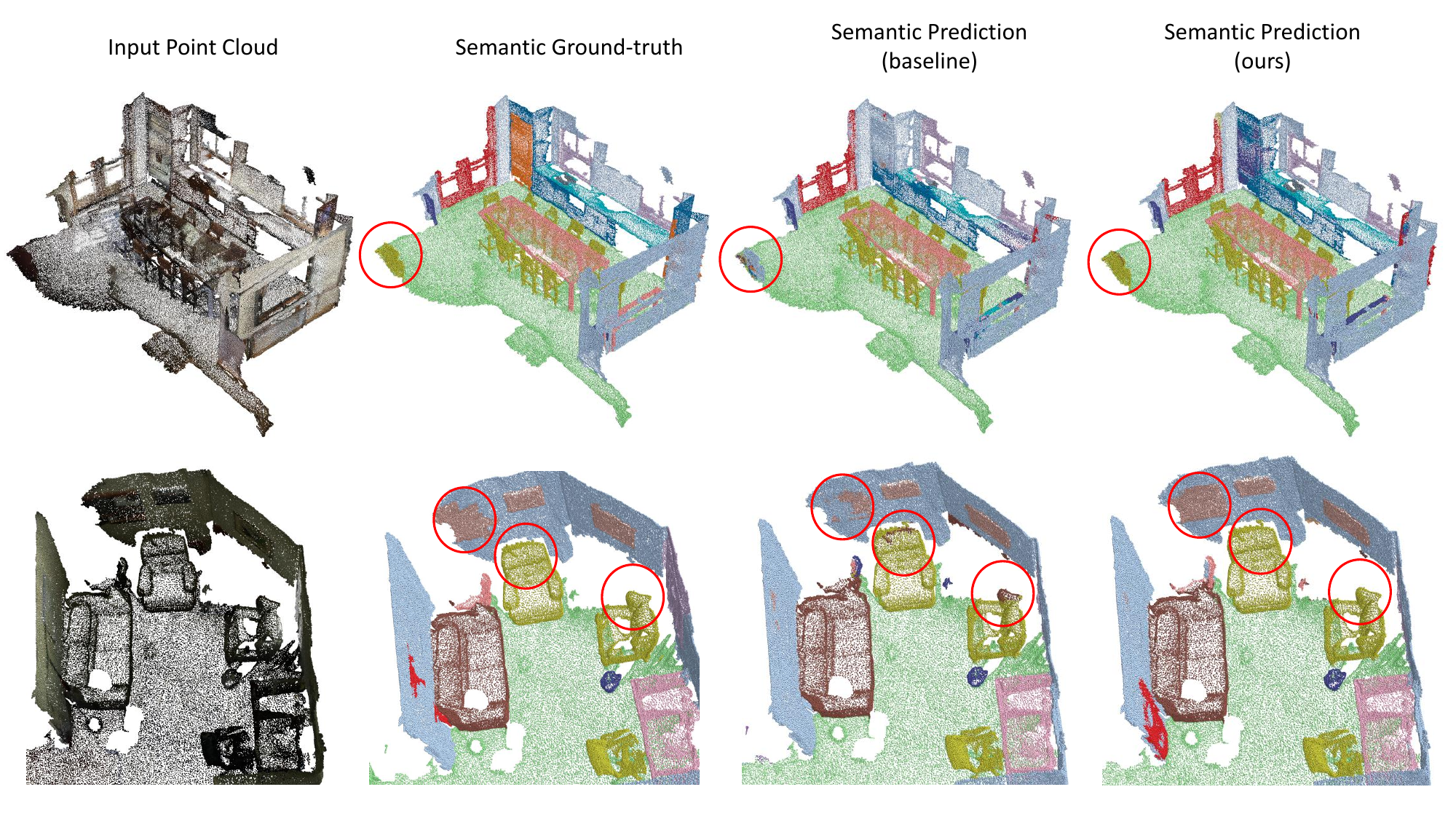}
	\end{center}
	\caption{Qualitative Results of 3D Semantic Segmentation on ScanNet-v2 \cite{dai2017scannet} validation set. We compare the semantic predictions of the network trained with additional synthetic virtual data from 3D-VirtFusion and its fully-supervised baseline.}
	\label{fig:3D_scannet}
\end{figure*}

\section{Experiment on 3D Semantic Segmentation}
To assess the effectiveness of our proposed data augmentation approach, we use the submanifold sparse convolution \cite{Submanifold} based U-Net structure from PointGroup \cite{pointgroup} as our backbone for 3D semantic segmentation task on ScanNet-v2 dataset \cite{dai2017scannet}. In this experiment, we combine both real data and synthetic data generated using our proposed method to train a network from scratch. We then evaluate this approach by comparing it with a model trained solely through fully supervised learning.

Following the same backbone parameters in \cite{pointgroup}, we use the voxel size of $2 cm$ and $7$ layers of U-Net. The batch size is set as $4$. The whole training process is on a single NVIDIA RTX 3090-ti GPU card, using Adam solver for optimization and an initial learning rate of $0.001$. We first create a diversified data element pool and then randomly stitch 9 elements into each virtual scene. The original training dataset comprises 1201 scenes, to which we add an additional 300 randomly generated virtual scenes to augment the training process. 

While data augmentation presents an intriguing phenomenon where augmenting the quantity of training data doesn't always yield a linear improvement in performance, it remains compelling to investigate whether augmented data can further enhance results on ScanNet. There are cases where increasing the diversity of data through augmentation can hardly improve performance, particularly when the original dataset lacks diversity. In such scenarios, data augmentation can help the model generalize better to unseen data and learn more robust representations. However, these improvements may not always be apparent when evaluating the model on a validation set.




\section{Discussions}
By employing a range of diverse augmentation strategies, our method enriches datasets, bolstering the training model's robustness, improving its generalization capabilities, and mitigating overfitting. One of the primary advantages of 3D-VirtFusion is its ability to leverage the vast knowledge encoded within large pre-trained models to generate high-quality augmented data. We can effectively expand the diversity of training datasets, thereby improving the model's ability to generalize to unseen data. This can prevent the model from memorizing specific patterns in the training data and encourage it to learn more robust and generalizable representations. 

Furthermore, 3D-VirtFusion offers a scalable and efficient approach to data augmentation, particularly in scenarios where collecting large volumes of labeled data is impractical or costly. By leveraging pre-trained large models, we can generate synthetic data quickly and cost-effectively, allowing us to overcome limitations associated with small or imbalanced training datasets.


\begin{figure*}[t]
	\begin{center}
		\includegraphics[width=0.9\linewidth]{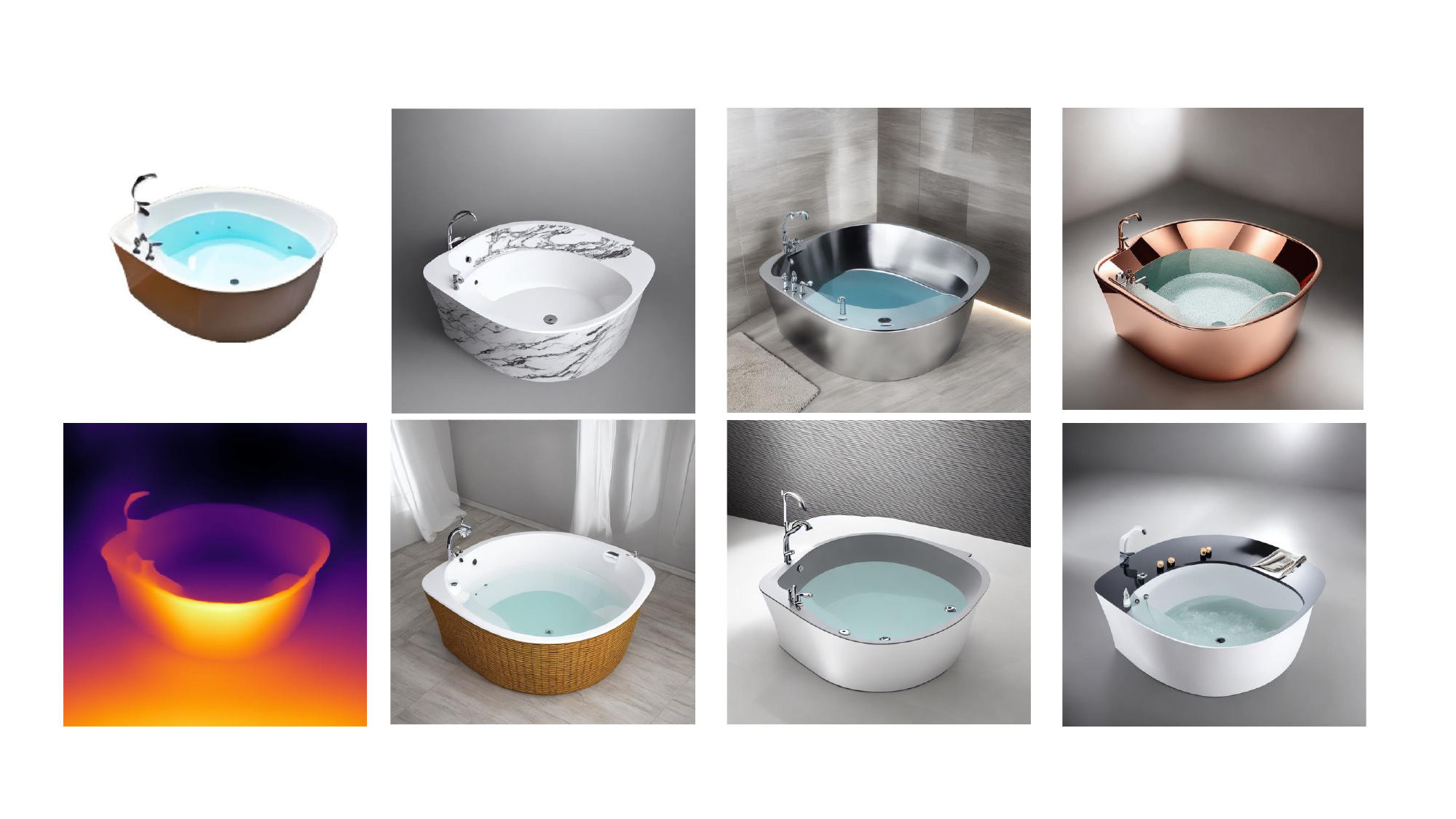}
	\end{center}
    \vspace{-8mm}
	\caption{Texture augmentation on a generated 2D object.}
	\label{fig:2D_texture_aug2}
\end{figure*}

\subsection{Challenges and Future work}
Despite the benefits, there are some challenges and considerations that warrant further investigation.

\subsubsection{Improving Data Quality from Generative Models}

One significant challenge in generative model is the potential for occlusion, partial visibility, and distortion of generated objects in both 2D and 3D settings. Despite efforts to mitigate these issues through text prompt design, such as centered, clean background, no occlusion, more effective techniques are needed to reduce the occurrence of occluded or distorted objects during data augmentation. Besides, certain augmentations from large foundation models may not effectively simulate real-world variations or may introduce unrealistic patterns into the data. In such cases, augmented samples may not contribute meaningfully to the model's learning process and could even degrade performance. Future work should focus on designing effective methods to identify and reject unacceptable generated images, ensuring that only high-quality data is used to train machine learning models.


\subsubsection{Domain Alignment and Fine-tuning}
Aligning the data domain of augmented data with that of the target dataset presents another challenge. While current approaches are generalized, there is a need to develop techniques for fine-tuning and matching augmented data to the specific domain of the target dataset. This is particularly crucial for tasks with distinct domain characteristics.

\subsubsection{Evaluation Metrics for Augmented Data}
Based on our review on existing literature, quantitatively evaluating the performance of point cloud data augmentation methods remains challenging. Existing evaluation methods often rely on downstream task metrics applied to specific datasets, lacking effective metrics for assessing the real quality and impact of augmented data.

\subsubsection{Theoretical Understanding and Frameworks}
A deeper theoretical understanding of data augmentation mechanisms is essential to support its design and implementation. While empirical studies have demonstrated efficacy, advancing theoretical understanding can further enhance effectiveness and applicability across domains.

Determining the amount of data to blend with the original data remains subjective and challenging. The optimal dataset size lacks theoretical guidance and is often determined empirically, tailored to specific tasks and models. Future research efforts should prioritize the development of standardized evaluation metrics capable of quantifying the diversity, fidelity, and effectiveness of augmented data.

\begin{figure*}[t]
	\begin{center}
		\includegraphics[width=0.9\linewidth]{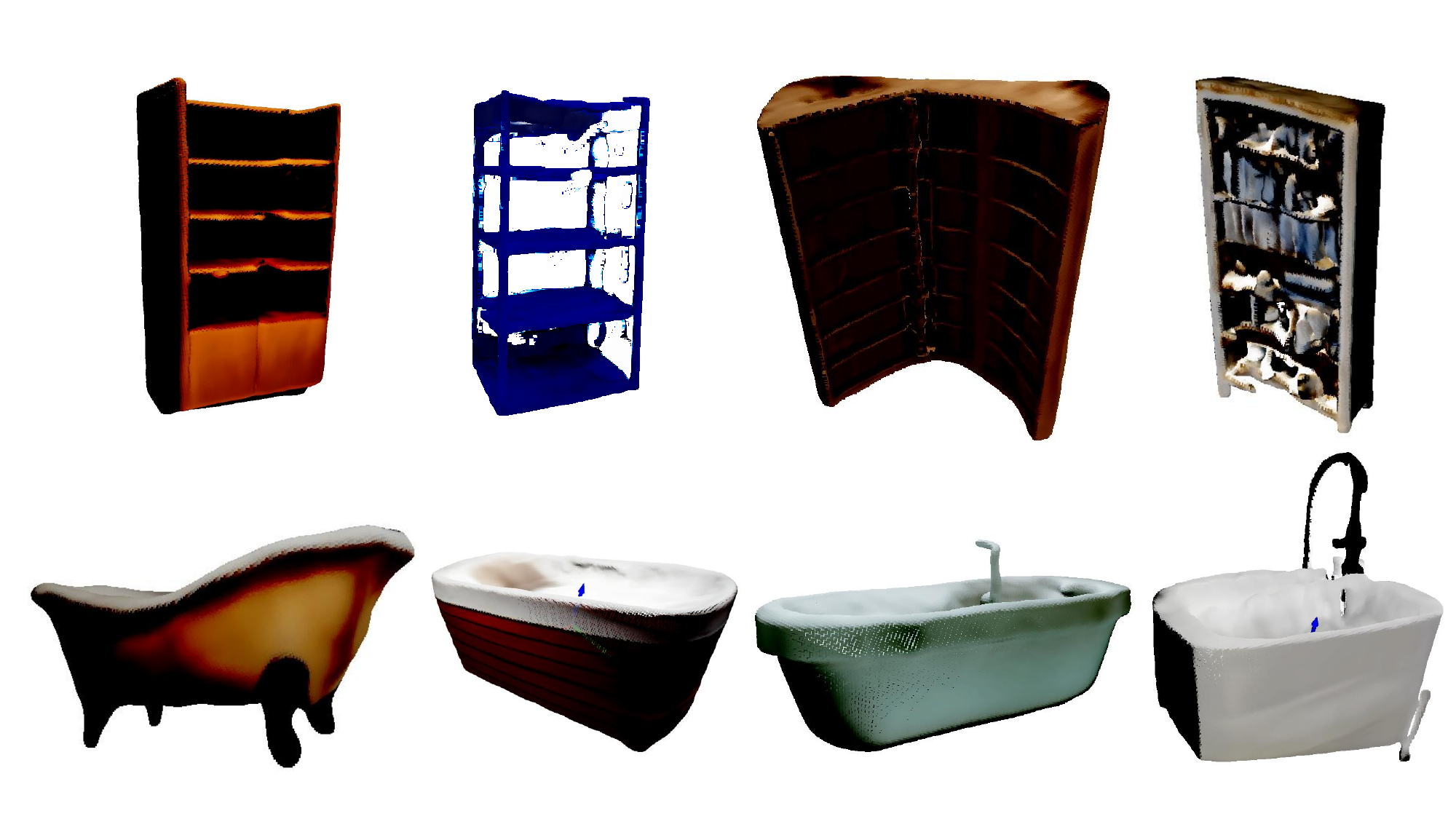}
	\end{center}
	\caption{Examples of generated 3D object.}
	\label{fig:3D_examples}
\end{figure*}

\subsection{Qualitative Results}
We show more qualitative results of generated object samples in Figure \ref{fig:2D_texture_aug2} and Figure \ref{fig:3D_examples}. The 3D semantic segmentation predictions on ScanNet \cite{dai2017scannet} validation set are shown in Figure \ref{fig:3D_scannet}.


{
    \small
    \bibliographystyle{ieeenat_fullname}
    \bibliography{supp}
}

%% file: preamble.tex
\usepackage[dvipsnames]{xcolor}


%% file: sec/0_abstract.tex
\begin{abstract}
 Data augmentation plays a crucial role in deep learning, enhancing the generalization and robustness of learning-based models. Standard approaches involve simple transformations like rotations and flips for generating extra data. However, these augmentations are limited by their initial dataset, lacking high-level diversity. Recently, large models such as language models and diffusion models have shown exceptional capabilities in perception and content generation. In this work, we propose a new paradigm to automatically generate 3D labeled training data by harnessing the power of pretrained large foundation models. For each target semantic class, we first generate 2D images of a single object in various structure and appearance via diffusion models and chatGPT generated text prompts. Beyond texture augmentation, we propose a method to automatically alter the shape of objects within 2D images. Subsequently, we transform these augmented images into 3D objects and construct virtual scenes by random composition. This method can automatically produce a substantial amount of 3D scene data without the need of real data, providing significant benefits in addressing few-shot learning challenges and mitigating long-tailed class imbalances. By providing a flexible augmentation approach, our work contributes to enhancing 3D data diversity and advancing model capabilities in scene understanding tasks.
\end{abstract}

%% file: sec/1_intro.tex
\section{Introduction}

The proposition of 3D virtual data generation stands as a pivotal necessity in contemporary research and application domains due to several compelling reasons. Primarily, the surge in demand for advanced 3D models across diverse industries, including computer vision, robotics, augmented reality, and virtual reality, underscores the importance of abundant and high-quality 3D data. 

In contrast to their 2D counterparts, 3D datasets offer a richer representation of the real world, encapsulating spatial information crucial for accurate scene understanding and interaction. However, despite the burgeoning need for 3D data, its availability remains significantly limited compared to 2D data. This scarcity can be attributed to various factors, prominently including the inherently complex nature of 3D data acquisition, processing, and annotation. Unlike 2D images, capturing 3D scenes necessitates sophisticated equipment, specialized expertise, and substantial time investment. Moreover, the manual annotation of 3D data is considerably more labor-intensive and challenging, exacerbating the scarcity issue. Consequently, the development of efficient and scalable methods for generating 3D virtual data emerges as an imperative solution to bridge this gap and facilitate advancements in 3D perception, modeling, and analysis. By automating the generation process and circumventing the constraints associated with real-world data collection, virtual data generation techniques offer the potential to democratize access to diverse and  voluminous 3D datasets. 


In deep learning tasks, achieving balanced class distributions within datasets is essential for ensuring the effectiveness and fairness of models. Class imbalance occurs when certain classes or categories of data are significantly underrepresented compared to others. This imbalance can lead to biased model predictions, where the minority classes are often overlooked or misclassified. Traditional methods for addressing class imbalance, such as oversampling or undersampling, may not be effective in scenarios where the distribution of classes is heavily skewed or there is very little data available for certain classes. Moreover, manual data collection and annotation efforts to mitigate class imbalances can be resource-intensive and time-consuming. In recent years, the advent of generative techniques for data augmentation, particularly in the context of 3D data, has provided a promising avenue for addressing class imbalance challenges. By generating synthetic 3D data, researchers can effectively augment minority class samples, thereby rebalancing the dataset and improving model performance. This approach not only mitigates the need for extensive manual data collection but also enables the creation of diverse and representative datasets that better reflect real-world scenarios. 

Generating high-quality 3D data has historically posed significant challenges due to the complexity and resource-intensive nature of data acquisition and annotation processes. Recently, pretrained large language and vision foundation models and AI-Generated Content (AIGC), have opened up new opportunities in this domain. These models, pretrained on vast amounts of 2D data, possess remarkable generalizability and imaginative capabilities. Over the last year, there has been substantial growth in research focused on large model-based data augmentation. However, the majority of these studies have focused on 2D data. Exploration into 3D data augmentation using generative models remains relatively underdeveloped.

In this study, we aim to address the challenge of limited labeled training data for 3D scene understanding tasks, without the need for explicitly collecting new data. We propose a novel approach that combines the strengths of text-to-image (T2I) diffusion models and ChatGPT-generated text prompts to generate synthetic images that accurately depict the structural descriptions provided in the text prompts. Additionally, we employ ControlNet to generate various appearance objects based on spatially-aligned conditions derived from depth map prediction and ChatGPT-generated texture descriptions as text prompts. These augmented images are further enhanced through automatic drag-based editing to introduce a greater diversity of objects. Finally, the 2D images are reconstructed into 3D objects and randomly composited into 3D synthetic virtual scenes. Notably, each individual object is generated based on the text prompt corresponding to a specific class. Thus the generated virtual scenes inherently possess semantic and instance labels, derived directly from the initial text prompts. These label information can be directly employed by downstream tasks, such 3D semantic segmentation \cite{3DSemanticSegmentationWithSubmanifoldSparseConvNet, Submanifold, dong2023leveraging}, 3D instance segmentation\cite{dong2022learning, dong2023collaborative,dong2023weakly}, and 3D object detection\cite{qi2019deep, Shi_2019_CVPR}.






Overall, our main contributions can be summarized as:

\begin{itemize}
    \item We introduce 3D-VirtFusion, an automatic augmentation pipeline based on various language and vision foundation models that generates 3D point cloud scenes without the need for input data. This off-the-shelf solution can enrich existing 3D datasets, thereby improving performance in scene understanding tasks.
    
    \item We propose a series of techniques aimed at enhancing the diversity of generated objects across structural, appearance, and textural perspectives. These techniques include chatGPT generated text prompt and automatic drag-based editing, facilitating the creation of a more diverse dataset. Such diversity is essential for training robust deep learning models to generalize well across different scenarios.
    
    \item We design a stitching algorithm that combines objects into 3D scenes with flexible templates. This algorithm incorporates random selection, rotation, and translation functionalities to facilitate flexible scene composition, contributing to more realistic and diverse virtual environments.
    

\end{itemize}


\begin{figure*}[t]
    \vspace{-8mm}
	\begin{center}
		\includegraphics[width=1.0\linewidth]{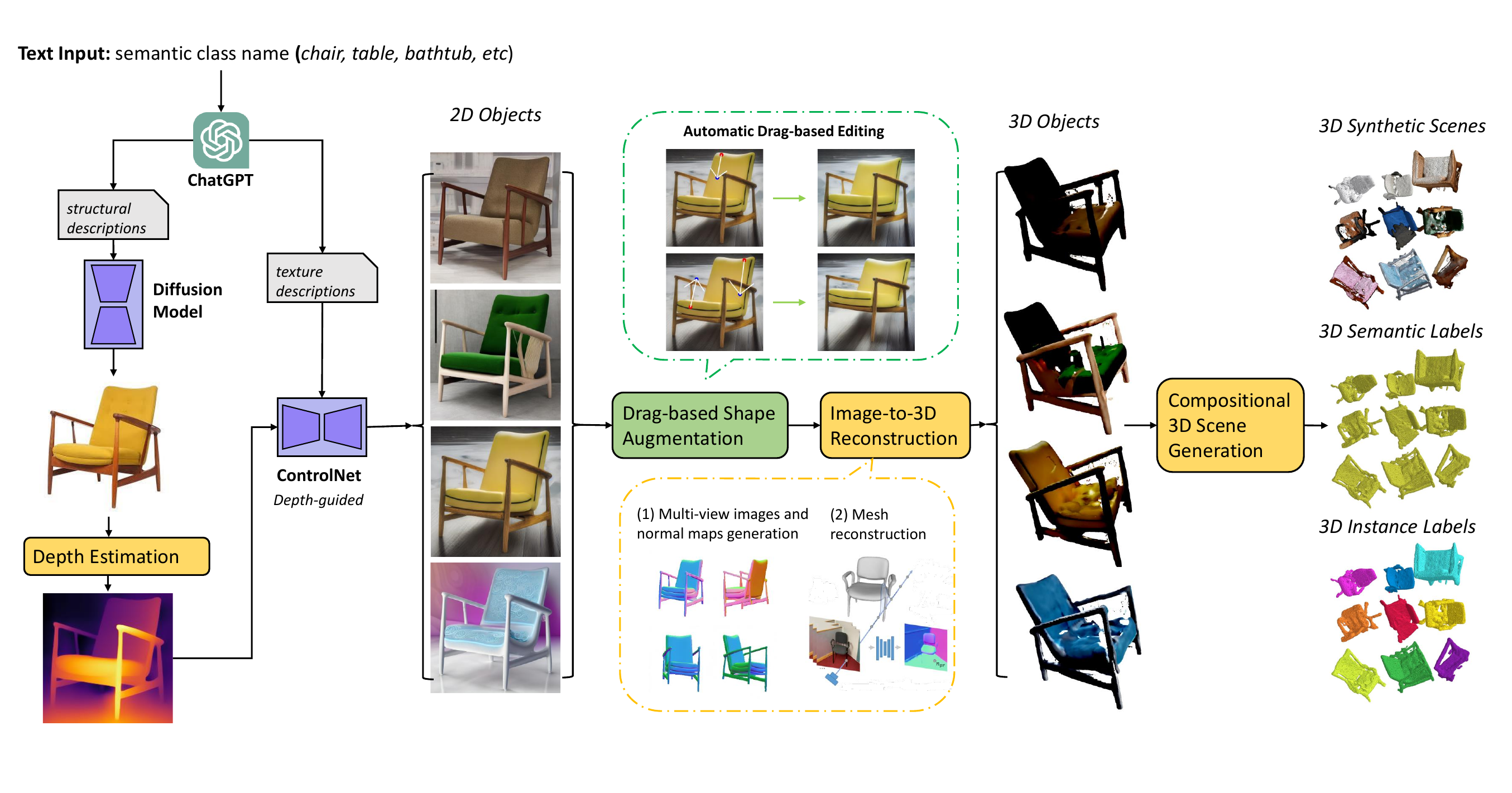}
	\end{center}
	\vspace{-8mm}
	\caption{Overview of proposed 3D-VirtFusion pipeline. Given a target semantic class, our method (3D-VirtFusion) consists of five steps: (1) Generate 2D object images via diffusion model \cite{StableDiffusion} and ChatGPT generated diversified structural descriptions as text prompt. (2) Produce depth map via Depth-Anything \cite{depthanything} and diversified texture descriptions via ChatGPT to guide ControlNet \cite{zhang2023adding_controlnet} in augmenting 2D objects into different appearances. (3) Employ proposed automatic drag-based shape augmentation method to further diversify data. (4) Adapt wonder3D \cite{long2023wonder3d} to make high-quality 3D reconstruction from each single images. (5) Utilize proposed template-based stitching algorithm to fuse augmented 3D objects into random 3D scenes, while simultaneously generating pixel-level semantic labels and instance labels.}
	\label{fig:Pipeline}
\end{figure*}

%% file: sec/2_relative.tex
\section{Related Work}
\subsection{Generative Data Augmentation on 2D images}
Generative models, such as VAEs \cite{VAE} and GANs \cite{GAN} have gained significant attention in recent years for generating photo realistic images. GANs \cite{GAN, BigGAN, VQ-VAE-2, StyleGANRender} are comprised of two neural networks, a generator and a discriminator, which are trained jointly. The generator learns to produce synthetic images that are indistinguishable from real ones, while the discriminator learns to differentiate between real and fake images. However, GANs can suffer from mode collapse and training instability. VAEs offer an alternative approach to generating synthetic data by learning a latent space representation of the input data. However, VAEs may struggle to generate high-quality images with fine-grained details. Diffusion models \cite{DDPM,ImprovedDDPM,Imagen,GLIDE,unCLIP} have gained more interest than GANs recently due to the ability to generate higher-quality samples. These models can be trained on large-scale datasets and demonstrate strong generalizability across diverse domains.

\subsection{Diffusion Model-Based Image Editing}
Diffusion model-based image editing tasks \cite{huang2024diffusion} enable the synthesis of visual content and can be generally classified into three main categories: semantic editing, stylistic editing, and structural editing. Beyond Text-to-Image (T2I) generation, more specific conditions are employed to enhance fidelity and ensure precise control. GLIGEN \cite{li2023gligen} allows for using grounding boxes as the condition for controllable image generation. SpaText \cite{Avrahami_2023_CVPR_spatext} and Make-A-Scene \cite{make_a_scene} propose to use semantic segmentation masks to guide image generation. Apart from segmentation maps, ControlNet \cite{zhang2023adding_controlnet} and T2I-Adapter \cite{mou2023t2i} can incorporate with various other input format such as depth map, canny edges, sketches as conditions. DragGAN \cite{pan2023draggan} and DragDiffusion \cite{shi2023dragdiffusion} provide flexible and precise controllability in a user-interactive manner, deforming shapes that consistently follow the object's rigidity.


\subsection{3D Generation from a Single Image}
Recently, the field of 3D generation has witnessed
rapid growth with the emergence of diffusion models and implicit neural representations such as NeRF \cite{mildenhall2020nerf} and Gaussian Splatting \cite{kerbl3Dgaussians}. Image-to-3D methods \cite{melaskyriazi2023realfusion, Magic123, shen2023anything3d, Tang_2023_ICCV, liu2023zero1to3, shi2023zero123plus, liu2023one2345, liu2023one2345++, tang2024lgm, long2023wonder3d, TriplanMeetsGaussian} typical follow a pipeline optimize a 3D neural representation via SDS loss \cite{Poole2022DreamFusionTU} and use neural rendering to generate multi-view images and reconstruct object into 3D space. 


\subsection{Data Augmentation on 3D Point Clouds}
Traditional methods such as PointAugment \cite{li2020pointaugment} and PointWOLF \cite{Kim_2021_ICCV}  apply geometric or statistical transformations to point cloud data, such as translation, rotation, scaling, noise addition, point removal, jittering, and density reduction, but may struggle to capture complex semantics. PointMixUp \cite{PointMixup} is designed to generate new examples via shortest-path interpolation functions. RSMix \cite{lee2021regularization} and SageMix \cite{lee2022sagemix} combines two point clouds into one continuous shape as augmented data. SageMix \cite{lee2022sagemix} adopt saliency-guided Mixup, which can preserve point clouds' salient local structures. Nevertheless, the augmented data generated by Mixup methods offer only marginal improvements and lack the capability to produce a diverse objects. TTA \cite{vu2023testtime} use point cloud upsampling with surface approximation as a test-time augmentation technique. PUGAN \cite{li2019pugan} empolys a GAN framework to upsample and augment point cloud data. The method is designed to complete small patch-level holes but has limited ability filling larger gaps in point clouds.  In contrast to methods that focus on single-object datasets, Mixed3D \cite{Mix3D_Nekrasov213DV} is introduced specifically for augmenting 3D scenes, achieving this by blending two scenes to create a new training sample. Existing 3D augmentation methods \cite{zhu2024advancements} face a common limitation: they can only augment data based on the provided real data, thus severely restricting the diversity of the augmented data. To this end, we propose a method that leverages the capabilities of foundational models to perform zero-shot augmentation that does solely rely on existing data.

%% file: sec/3_method.tex
\section{Method}
Our methodology is outlined in Figure \ref{fig:Pipeline} to provide a comprehensive overview.  The process commences with the generation of diversified 2D images of single objects using diffusion models and ChatGPT-generated text prompts. Subsequently, we automatically adjust the shapes of objects within these 2D images. Following this, the augmented 2D images are transformed into 3D objects. These 3D objects are then randomly composed to construct synthetic virtual scenes. Notably, the generated virtual scenes are equipped with semantic and instance labels, facilitating downstream task training.



\begin{figure}[t]
	\begin{center}
		\includegraphics[width=1.0\linewidth]{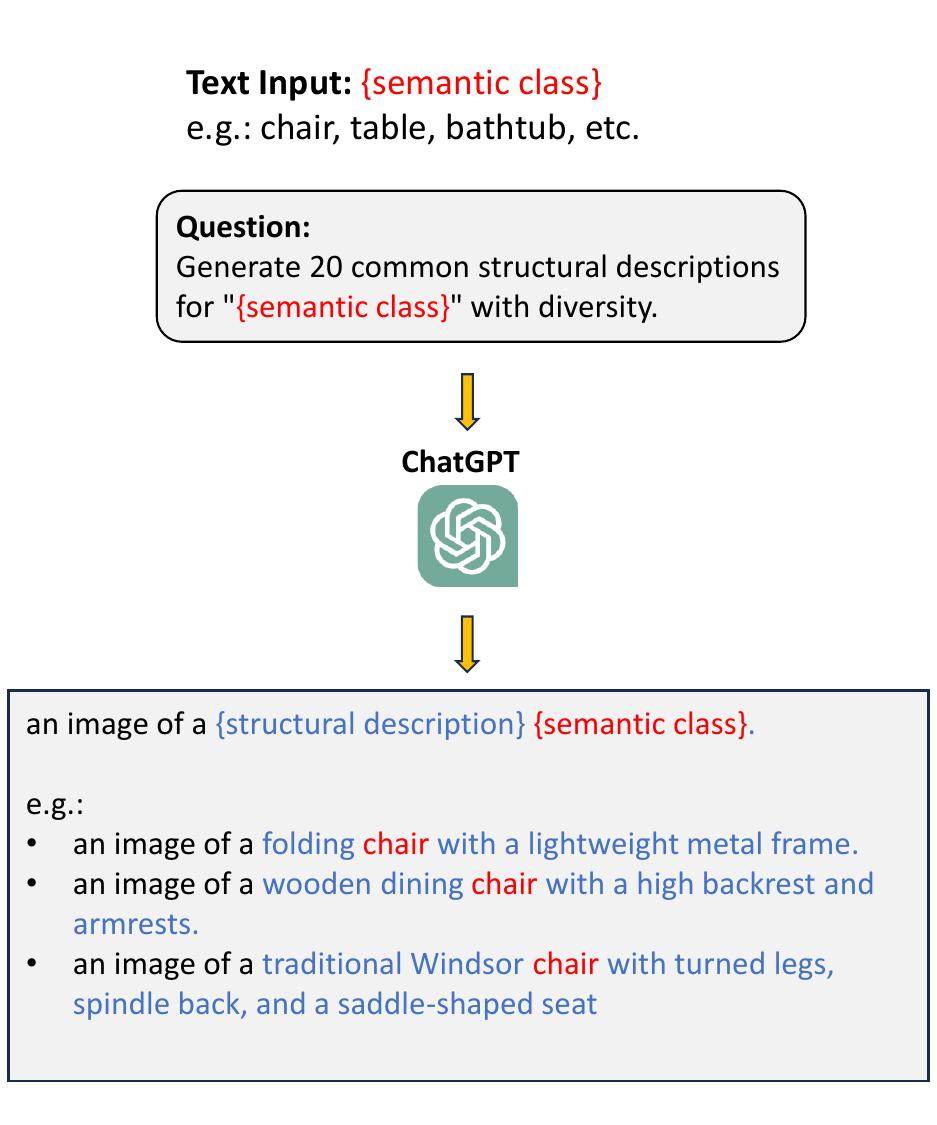}
	\end{center}
	\vspace{-4mm}
	\caption{Generation of structural descriptions with ChatGPT. When provided with a target semantic class, we utilize a template to pose a question to ChatGPT, prompting it to generate diverse structural text prompts. These prompts are then employed to facilitate image generation with the diffusion models.}
	\label{fig:strucural_prompt_gen}
\end{figure}

\subsection{2D Image Generation with Diffusion Models}

Diffusion models \cite{OriginalDiffusionPaper, DDPM, ImprovedDDPM, DDIM, StableDiffusion} are probabilistic generative models used primarily for image generation tasks. Inspired by thermodynamic diffusion \cite{OriginalDiffusionPaper}, they generate samples by iteratively adding Gaussian noise to an initial noise distribution $p(x_T) = \mathcal{N}(x_T; 0, I)$ until it converges. This process is a Markov chain with learned Gaussian transitions:

\begin{equation}
    p_{\theta}(x_{0:T}) = p(x_T) \prod_{t = 1}^{T} p_{\theta} (x_{t - 1} | x_t)
\end{equation}

Transitions $p_{\theta} (x_{t - 1} | x_t)$ in diffusion model are craft to decrease variance over time, following a predefined schedule denote by $\beta_1, \hdots, \beta_T$. This gradual reduction aims to ensure that the final sample $x_0$ reflects a representation of the true distribution. These transitions are defined via a fixed covariance $\Sigma_t = \beta_t I$ and a learned mean $\mu_{\theta} (x_t, t)$ defined below:

\begin{equation}\label{eqn:reverse-step}
    \mu_{\theta} (x_t, t) = \frac{1}{\sqrt{\alpha_t}} \left( x_t - \frac{\beta_t}{\sqrt{1 - \tilde{\alpha}_t}} \epsilon_{\theta} ( x_t, t ) \right)
\end{equation}

As outlined in \cite{DDPM}, the parameterization is based on the optimization of the reverse process. In this context, $\epsilon_{\theta} (\cdot)$ represents a neural network trained to process noisy input $x_t$ and predict the noise added to it. Based on samples $x_0$ and noise $\epsilon \sim \mathcal{N}(0, I)$, we can compute $x_t$ at any given timestep using the following equation:

\begin{equation}
    x_t (x_0, \epsilon) = \sqrt{\tilde{\alpha}_t} x_0 + \sqrt{1 - \tilde{\alpha}_t} \epsilon
\end{equation}

Here, $\alpha_t = 1 - \beta_t$ and $\tilde{\alpha}_t = \prod_{s = 1}^{t} \alpha_t$ are defined based on the schedule $\beta_T$. In this work, we use pretrained Stable Diffusion models to perform text-to-image generation.


In order to obtain a diverse set of high-quality samples from diffusion models, we integrate ChatGPT into our framework to generate text prompts using a predefined template. As described in Figure \ref{fig:strucural_prompt_gen}, we insert the name of the target semantic class into our question template, prompting ChatGPT to produce multiple common structural descriptions. Leveraging the capabilities of ChatGPT allows us to acquire responses detailing the common structural patterns and typical architectural styles associated with the target class.


\begin{figure}[t]
	\begin{center}
		\includegraphics[width=1.0\linewidth]{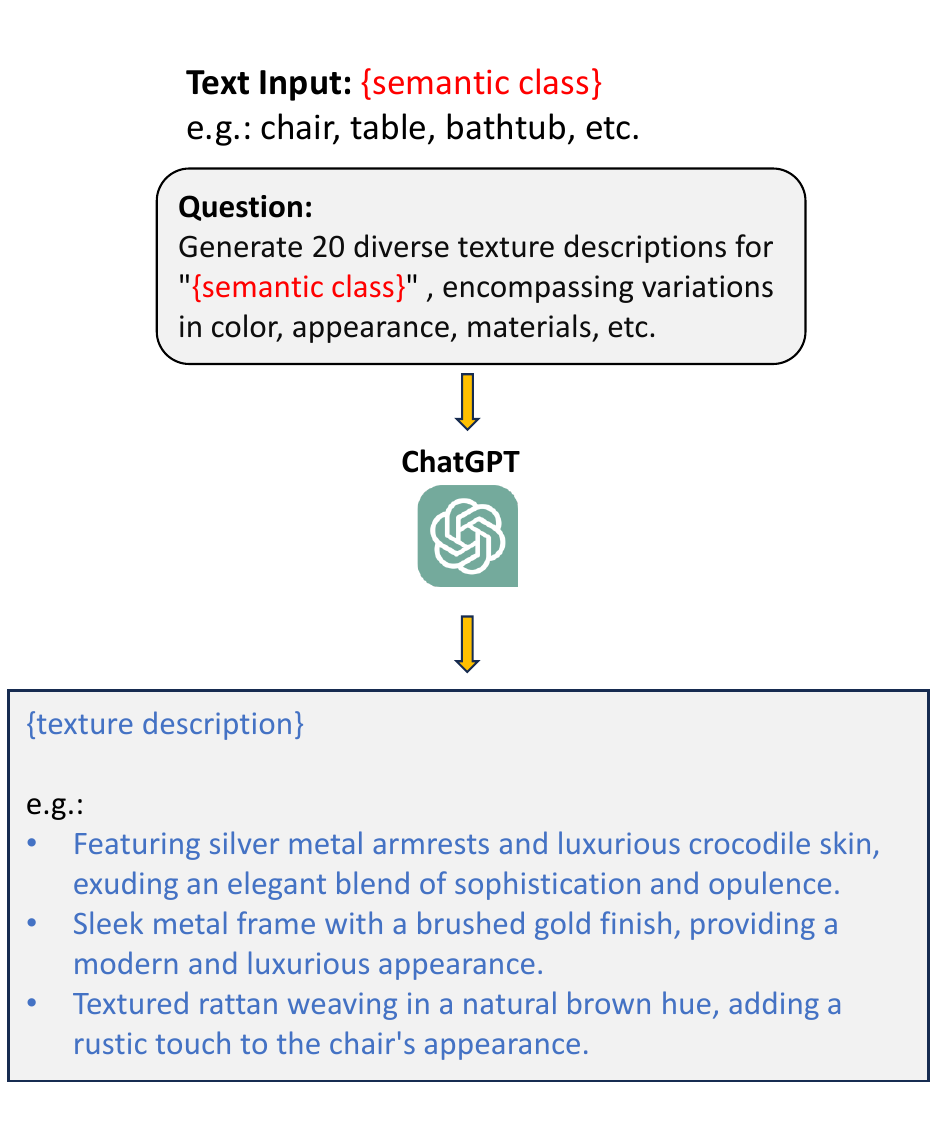}
	\end{center}
	\vspace{-4mm}
	\caption{Generation of texture descriptions with ChatGPT. When provided with a target semantic class, we utilize a template to pose a question to ChatGPT, prompting it to generate diverse texture text prompts. These prompts are then employed to facilitate image augmentation with ControlNet.}
	\label{fig:texture_prompt_gen}
\end{figure}

\subsection{2D Image Augmentation with ControlNet}

After generating a 2D object using the stable diffusion model, our objective is to produce diverse variations of the object, each with distinct appearance and texture. To accomplish this, we first conduct depth prediction on the generated image and utilize the resulting depth information as a condition for ControlNet \cite{zhang2023adding_controlnet}. ControlNet is a neural network architecture designed to incorporate spatial conditioning controls into existing pretrained diffusion models. It The recent foundation model, Depth Anything \cite{depthanything}, is employed to generate reliable monocular depth estimations. Concurrently, we employ ChatGPT to generate numbers of texture descriptions, which serve as text prompts for ControlNet, as shown in Figure \ref{fig:texture_prompt_gen}. By combining these inputs, the depth-guided ControlNet can generate varied versions of the object while preserving its fundamental structure (Figure \ref{fig:2D_texture_aug}). 


\begin{figure}[t]
	\begin{center}
        \vspace{-4mm}
		\includegraphics[width=0.85\linewidth]{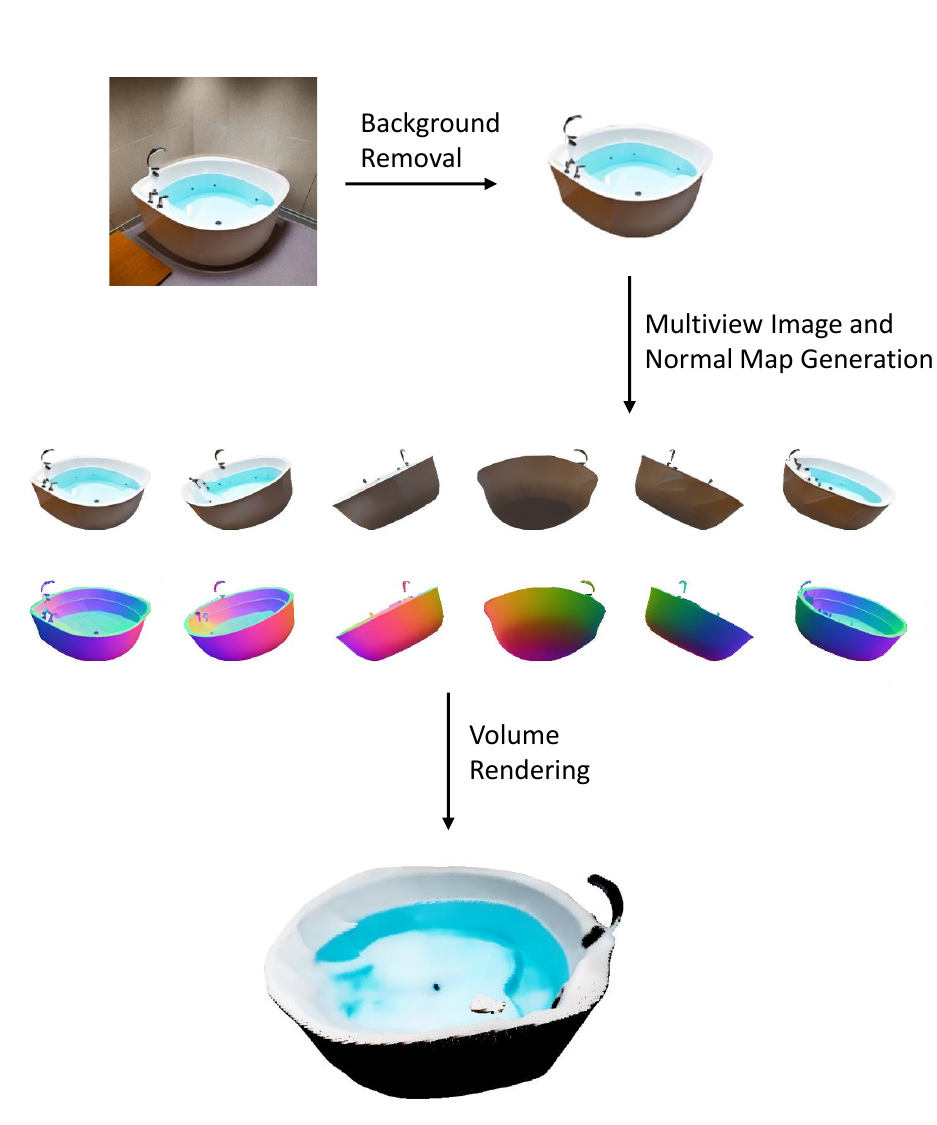}
	\end{center}
	\vspace{-4mm}
	\caption{3D Object Reconstruction Process}
	\label{fig:3d_recon}
\end{figure}

\subsection{Automatic Drag-based Shape Augmentation}

To further enhance the diversity of objects, we aim to make slight random adjustment to the shape of each textured object generated from previous steps. However, traditional 2D rigid transformations, such as resizing and rotation, come with several limitations. Firstly, they are restricted to modifying the size and orientation of objects in a fixed, predefined manner, which may not adequately capture the complex variations present in real-world scenarios. Additionally, these transformations often do not account for non-linear deformations or subtle changes in object shape, limiting their ability to accurately represent diverse object configurations. Moreover, traditional rigid transformations may introduce artifacts or distortions and lack the ability to incorporate 3D spatial information. Overall, these limitations hinder the capacity of traditional 2D rigid transformations to effectively model the full range of variability observed in real-world objects and scenes.  Recent advancements, such as DragGAN \cite{pan2023draggan} and DragDiffusion \cite{shi2023dragdiffusion}, leverage generative models for interactive control of shape manipulation. These models can accurately capture complex variations in object shapes and textures, enabling realistic and high-fidelity deformations.

Building upon this inspiration, we propose a novel strategy for automatic random drag-based shape augmentation. This approach eliminates the need for human interaction by introducing a controlled randomness to adjust the shape of textured objects. The implementation involves several steps. Firstly, we must train a Low Rank Adaptation (LoRA) model \cite{hu2022lora} using input images to facilitate rapid fine-tuning. LoRA entails preserving the original weights of the model while introducing trainable rank decomposition matrices into each layer. This process contributes to the production of higher-quality images. Afterwards, we randomly select one or two points on the object as seeding points, serving as the starting points for the shape adjustment process. Then, a random direction is chosen to determine the direction of the deformation. Following this, a target point along the chosen direction is selected. The distance between the seeding point and the target point follows a Gaussian normal distribution, relying on two key parameters: the mean $\mu$ and variance $\sigma^2$. This allows us to control the extent of the shape augmentation, ensuring robust and diverse deformations across different instances of object manipulation.





\begin{figure}[t]
	\begin{center}
		\includegraphics[width=1.0\linewidth]{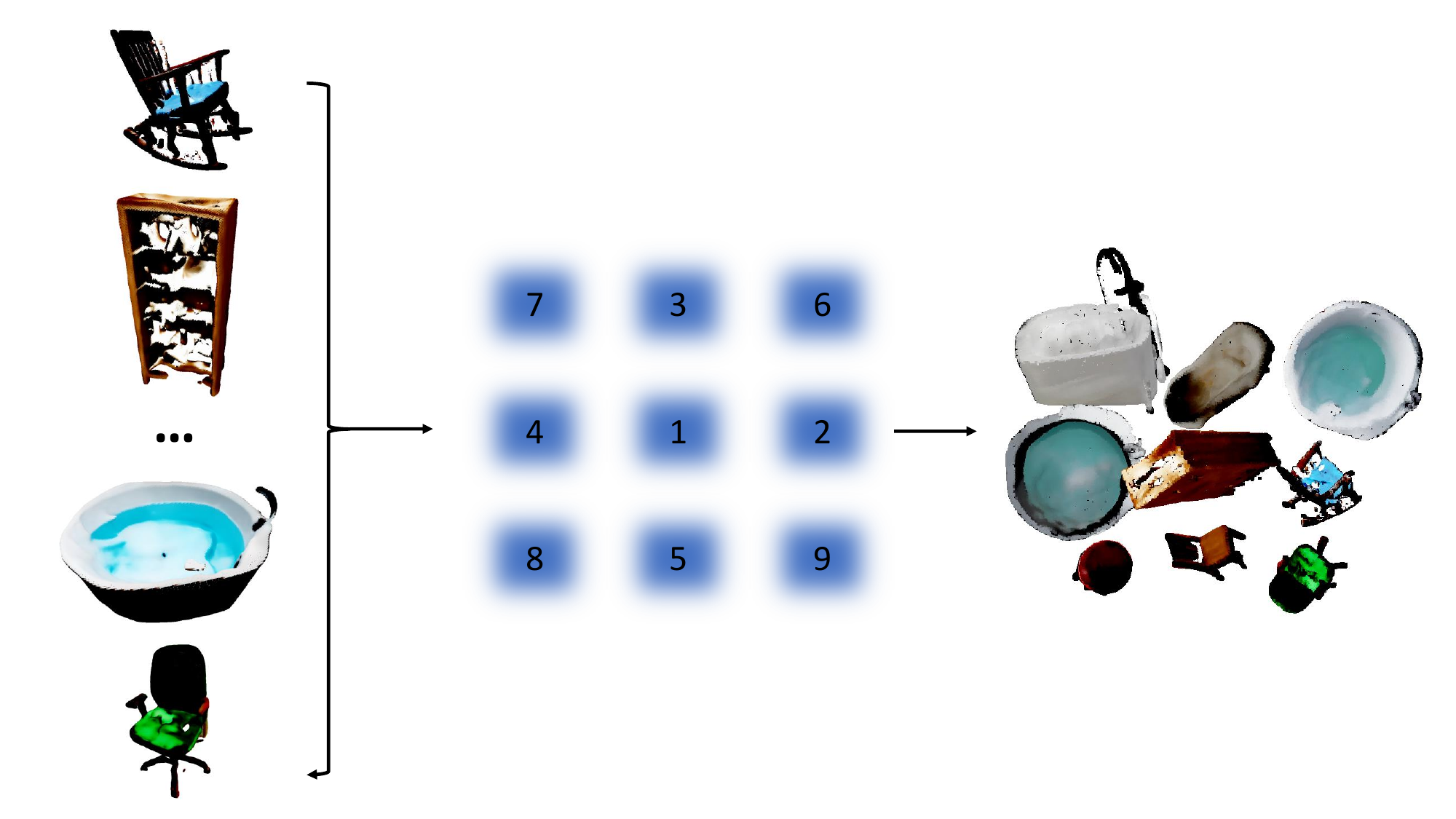}
	\end{center}
	\vspace{-4mm}
	\caption{Compositional 3D Scene Generation Process. Objects are sequentially stitched into the bird-view template following the location IDs.}
	\label{fig:strucural_scene_fusion}
\end{figure}

\subsection{Image-to-3D Reconstruction}
Single image to 3D reconstruction is a fundamental task in computer vision, aiming to infer the 3D geometry and structure of an object or a scene from a single 2D image. It involves the challenging process of recovering the depth, shape, and spatial layout of objects represented in the image, without any additional views or prior knowledge. In our approach, we aim to transform augmented 2D objects from previous steps into 3D space in high quality. To account for potential inaccuracies in the size of reconstructed 3D objects, we robustly adjust their dimensions based on their semantic class. As shown in Figure \ref{fig:3d_recon}, we enhance the quality of reconstruction by first removing the background in 2D images. Following Wonder3D \cite{long2023wonder3d}, we use a cross-domain diffusion model to generate consistent multi-view images with their corresponding normal maps. These normal maps serve as supervision for training a NeuRIS \cite{wang2021neus} network structured with Multi-layer Perceptrons (MLPs), implicitly encoding the 3D object. Subsequently, we extract a 3D mesh from the trained density field.


\begin{algorithm} [t]
\caption{Compositional 3D Scene Generation}
\begin{algorithmic}[1]
\State \textbf{Input:} 3D object sample sets $O = \{O_1, O_2, ..., O_n\}$ 
\State \hspace{0.75cm}  semantic labels $O_{sem} = \{{O_1}_{sem}, {O_2}_{sem}, ..., {O_n}_{sem}\}$ 
\State \hspace{0.75cm}  number of objects in the scene $k$
\State \hspace{0.75cm}  point number threshold $\tau$

\State \textbf{Output:} 3D compositional scene $S$, semantic labels $S_{sem}$, 
\State  \hspace{0.75cm}  instance labels $S_{ins}$

\State Initialize empty scene $S \leftarrow \emptyset$

\For{$i = 1$ to $k$} \Comment{Number of objects in template}
    \State Randomly pick an object $o$
    \State Place object $o$ at the position ID $i$ on the template
    \State Randomly rotate object $o$
    \If{target position is occupied by existing objects}
        \State Shift object $o$ along the predefined direction
    \EndIf
    \State $S_{ins} \gets i$ 
\EndFor

\If{$N > \tau$} \Comment{Check if point number is beyond threshold}
    \State Randomly downsample $\tau$ points to be $S'$
    \State  $S \gets S'$, $S_{sem} \gets S_{sem}'$, $S_{ins} \gets S_{ins}'$   
\EndIf
    
\State Randomly rotate the entire scene $S$
\State \textbf{return} $S$, $S_{sem}$, $S_{ins}$

\end{algorithmic}
\label{algo:3D-VirtFusion}
\end{algorithm}

\subsection{Compositional 3D Scene Generation}
The subsequent step involves integrating the augmented 3D objects into scenes as part of the preparation for model training in downstream tasks. To accomplish this, we propose to use a template that can hold nine objects, as illustrated in Figure \ref{fig:strucural_scene_fusion}. The template is designed on the bird-view of objects. Considering the varying sizes of these objects, we establish flexible guidelines aimed at preventing overlap between objects. The objects are sequentially stitched into the template, one after the other. Oversized objects can cause the next object to be shifted aside. Additionally, we introduce randomness in both object-level and scene-level rotations, to improve the generalizability. The detailed process is explained in Algorithm \ref{algo:3D-VirtFusion}.

%% file: sec/4_experiments.tex
\begin{figure*}[t]
    \vspace{-4mm}
	\begin{center}
		\includegraphics[width=1.0\linewidth]{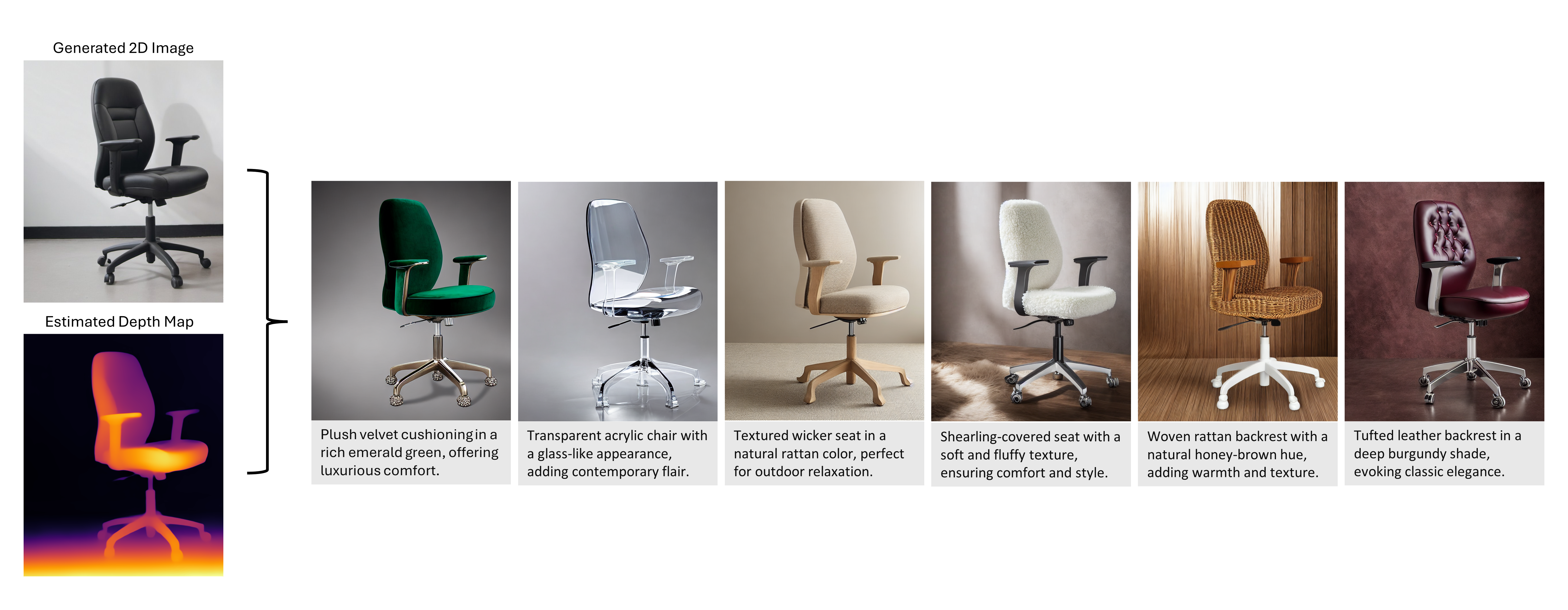}
	\end{center}
    \vspace{-8mm}
	\caption{Texture augmentation on a generated 2D object.}
	\label{fig:2D_texture_aug}
\end{figure*}

\begin{figure*}[t]
    \vspace{-4mm}
	\begin{center}
		\includegraphics[width=1.0 \linewidth]{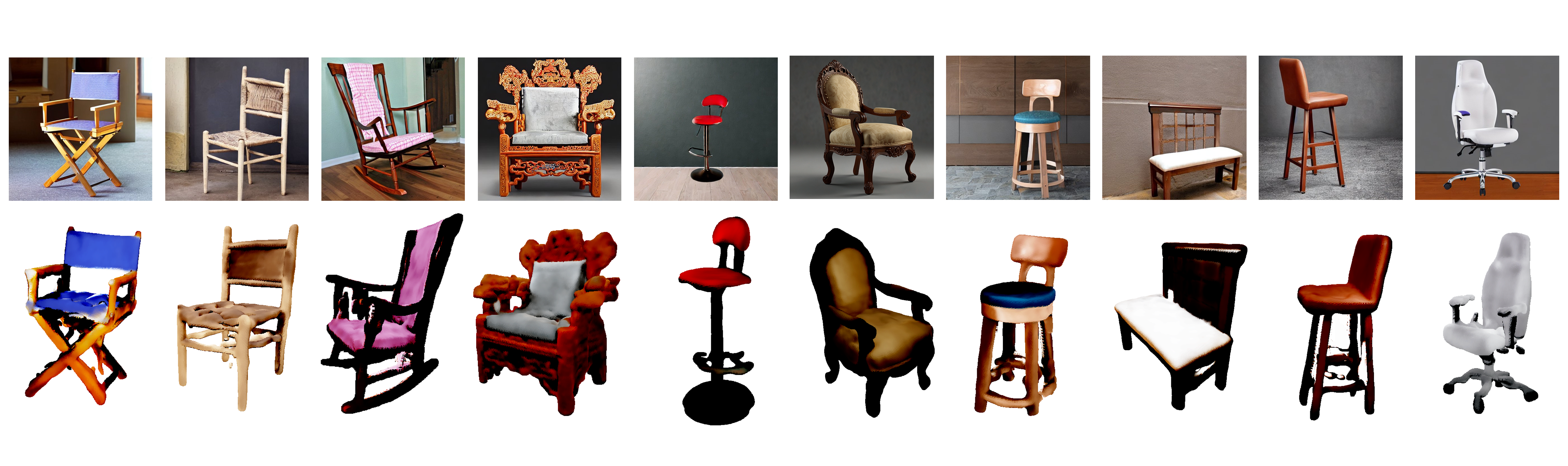}
	\end{center}
    \vspace{-8mm}
	\caption{Generated 2D objects and their corresponding reconstructed textured 3D meshes by wonder3D \cite{long2023wonder3d}. The objects exhibit high diversity, facilitated by our automated text prompt generation using ChatGPT.}
	\label{fig:3D_chairs}
\end{figure*}

\section{Experiments}
\subsection{Implementation Details}
We implement our proposed method to generate complementary data with various semantic classes, especially for indoor scenes. User
Unless somewhere specified, the image and point cloud results are produced with Diffusion model v1.5 \cite{StableDiffusion}, ChatGPT3.5, DepthAnything \cite{depthanything}, ControlNet \cite{zhang2023adding_controlnet}, DragDiffusion \cite{shi2023dragdiffusion} and wonder3D \cite{long2023wonder3d}. In our compositional 3D Scene Generation algorithm \ref{algo:3D-VirtFusion}, number of objects in template $k$ is set to 9 and point number threshold $\tau$ is 200k. We train the LoRA \cite{hu2022lora} of Dragdiffusion \cite{shi2023dragdiffusion} on a single GeForce RTX 3090 GPU with 24GB memory.

We adopt the ScanNet-v2 dataset \cite{dai2017scannet} as our target dataset. Containing 2.5 million RGB-D views across 1513 real-world indoor scenes, this dataset provides detailed semantic and instance labeling across 20 diverse object categories. In comparison to other 3D indoor datasets, ScanNet-v2 is distinguished by its comprehensiveness and widespread recognition.

\subsection{Qualitative Results}

\subsubsection{Texture Augmentation on 2D objects}
Figure \ref{fig:2D_texture_aug} illustrates examples of our generated images generated using text prompts describing textures. We initiate the process by using a generated 2D image as input. Subsequently, we employ Depth-Anything \cite{depthanything} to predict its depth map. ControlNet \cite{zhang2023adding_controlnet} utilizes this depth map as guidance and incorporates text prompts as instructions for augmented image generation. The 2D images produced by our method have
diversified appearance and effectively preserve the structure from the input image.

\subsubsection{Overall Data Generation Quality}
\begin{figure}[t]
	\begin{center}
		\includegraphics[width=1.0\linewidth]{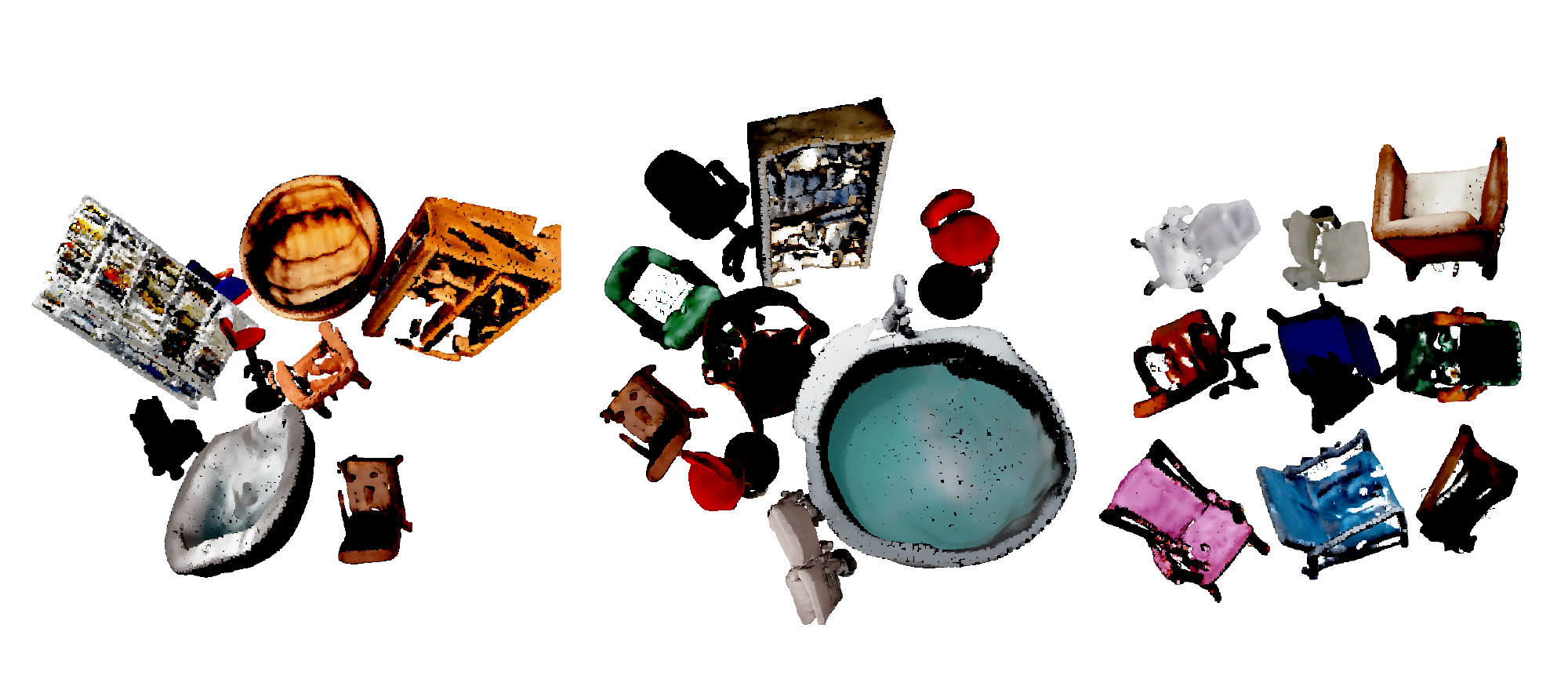}
	\end{center}
    \vspace{-8mm}
	\caption{Generated 3D Scenes by 3D-VirtFusion.}
	\label{fig:3D_scenes}
\end{figure}

Figure \ref{fig:3D_chairs} shows the generated 2D objects and their corresponding reconstructed textured 3D meshes using Wonder3D \cite{long2023wonder3d}. Given input semantic class of ``chair" and auto generated text prompts from ChatGPT, our method is capable of creating numerous objects under different styles and appearance. Figure \ref{fig:3D_scenes} shows the random synthetic 3D scenes with various generated 3D objects. The size of 3D objects has been regularized based on their semantic class. Our method is capable to generate unlimited 3D scenes from a pool of generated single objects at no extra cost. This capability facilitates robust training for downstream tasks such as 3D semantic segmentation and instance segmentation. There are generally two ways to utilize these synthetic 3D data: (1) conducting pretraining exclusively on our virtual dataset followed by fine-tuning on the target dataset, and (2) blending the synthetic data with the target dataset for joint training.

\subsubsection{Image-to-3D Reconstruction}
\begin{figure*}[t]
	\begin{center}
		\includegraphics[width=0.85\linewidth]{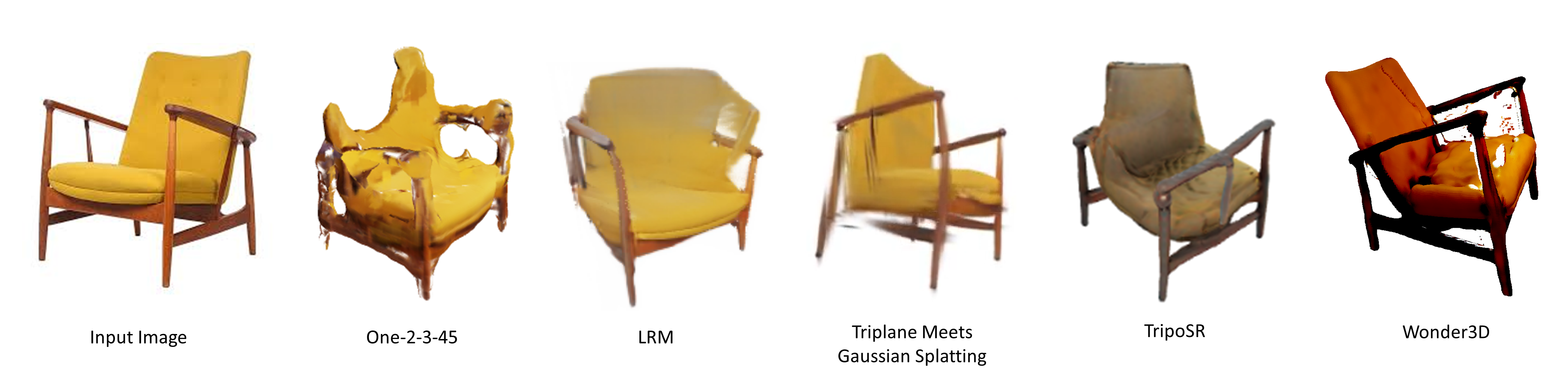}
	\end{center}
    \vspace{-8mm}
	\caption{Comparison of Image-to-3D Generation Methods.}
	\label{fig:Image_to_3D_comparison}
\end{figure*}

In Figure \ref{fig:Image_to_3D_comparison}, we evaluate different Image-to-3D methods \cite{liu2023one2345, TriplanMeetsGaussian, hong2023lrm, openlrm, long2023wonder3d} in our experiments. Our comparison reveals that Wonder3D \cite{long2023wonder3d} exhibits the highest reconstruction quality, characterized by minimal structural distortion or collapse. Method like Zero123 \cite{liu2023zero1to3} and One-2-3-45 \cite{liu2023one2345} can produce reasonable images but lack of multi-view consistency, which may lead to inconsistent 3D reconstruction results.  One-2-3-45++ \cite{liu2023one2345++} by SUDOAI and TRIPO by Sensetime can also produce high-quality image-guided 3D generation. However, being commercial products, their source code is not publicly available.

\subsection{Quantitative Results}
In Table \ref{tab:semantic_segmentation_scannet_val}, we present a comparison of the 3D semantic segmentation results obtained with and without the incorporation of synthetic virtual scene data from our augmentation method. The baseline results are assessed using PointGroup \cite{pointgroup} 's publicly released pretrained model, which is trained from scratch. Overall, our experiments show that incorporating synthetic data generated by our proposed 3D-VirtFusion can further improve the performance of our trained model by $2.7\%$ mIoU across 20 classes.

\begin{figure}
	\begin{center}
		\includegraphics[width=0.8\linewidth]{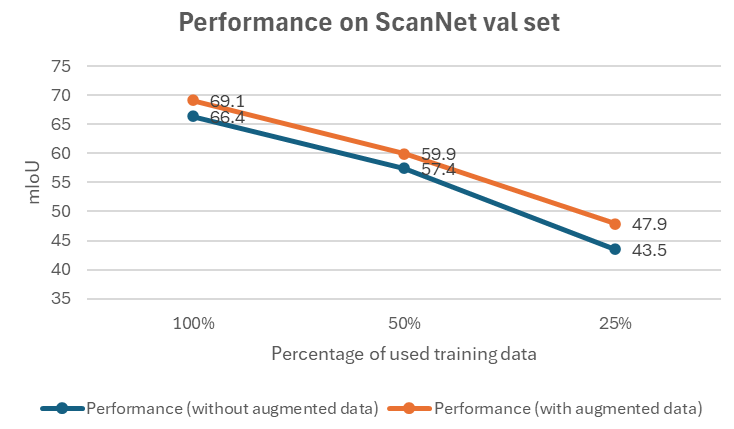}
	\end{center}
    \vspace{-4mm}
	\caption{Downstream Task Qualitative Comparison.}
	\label{fig:quantitative}
\end{figure}

In Figure \ref{fig:quantitative}, we assess the effectiveness of our augmented 3D data across different scenarios by varying the percentage of original data used for joint training (100\%, 50\% and 25\%). The results demonstrate that our augmented data are particularly beneficial when the original dataset is limited.

\begin{table*} [t]
    \resizebox{\textwidth}{!}{
    \begin{tabular}{c|cccccccccccccccccccc|c}
    \toprule
    \textbf{mIoU} & wall & floor & cab & bed & chair & sofa & tabl & door & wind & bkshf & pic & cntr & desk & curt & fridg & showr & toil & sink & bath & ofurn & \textbf{avg}\\
    \midrule
    Baseline&    81.4&94.7&59.1&77.8&87.3&76.9&67.6&52.4&58.7&78.2&27.4&57.7&60.8&65.4&41.2&60.4&87.4&58.9&82.6&52.8&\textbf{66.4}\\
    \midrule
    Ours&    
    83.7&94.8&64.6&79.4&88.4&81.0&69.5&58.4&62.1&77.5&31.2&60.4&61.2&68.3&45.4&65.9&90.7&61.9&81.7&56.4&\textbf{69.1} \textcolor{blue}{(+2.7)}\\
    \bottomrule
    \end{tabular}
    }
    \vspace{-2mm}
    \caption{Semantic segmentation results on ScanNet v2~\cite{dai2017scannet} validation set.}
    \label{tab:semantic_segmentation_scannet_val}
\end{table*}

\subsection{Discussions}


Our findings highlight the versatility of synthetic data augmentation in addressing key challenges faced in 3D computer vision, such as limited data availability, domain shift, and class imbalance. By leveraging synthetic data, we have been able to overcome these challenges and achieve improved performance on tasks including 3D object detection, semantic segmentation, and instance segmentation.

One of the key advantages of synthetic data augmentation with large foundation models is its ability to generate large amounts of diverse and annotated data at minimal cost. For our proposed method, we are able to generate $C \times N \times M \times P$ numbers of different 3D objects, where $C$ is the number of semantic class, $N$ is the number of initial generated objects in different structure, $M$ is the number of texture augmented samples for each of its input, $P$ is the number of Drag-based shape augmented samples for each of its input. Based on the large amount of generated 3D object, we can further construct unlimited random virtual scenes. This has significant implications for both research and industry applications, where access to labeled data is often limited or costly to acquire.

Furthermore, synthetic data augmentation can complement real-world datasets, providing a valuable source of additional training examples without introducing significant biases. By combining synthetic and real data in a joint training framework, we have observed further improvements in model performance, underscoring the potential for hybrid approaches to achieve state-of-the-art results in 3D computer vision tasks.

However, it is important to acknowledge the limitations of synthetic data augmentation. While synthetic data can simulate a wide range of scenarios, it may not fully capture the complexity and variability of real-world data. Therefore, careful consideration must be given to the design of synthetic datasets and the fidelity of the generated data to ensure that models trained on synthetic data generalize well to real-world environments.

Class ambiguity remains a challenge, particularly in datasets where class labels have multiple meanings. The distinction between certain classes, such as chairs and sofas, or tables and desks, can sometimes lack a clear boundary. Future research directions could explore methods for mitigating class ambiguity in synthetic data generation, such as incorporating context-aware labeling schemes or developing algorithms for disambiguating class labels based on image content.






%% file: sec/5_conclusion.tex
\section{Conclusion}

In this study, we have presented a comprehensive automatic synthetic data augmentation pipeline to address the challenge of limited labeled training data for 3D scene understanding tasks. Our proposed method, 3D-VirtFusion, leverages various large language and vision foundation models to generate diversified realistic 3D scenes with accurate pixel-level annotations, thus eliminating the need for human effort in 3D data collection and annotation. We propose techniques to enhance the diversity of generated objects across structural, appearance, and textural perspectives.  Overall, our work opens up new avenues for generating high-quality 3D virtual data for augmentation and aims to inspire further in-depth exploration in this direction.

%% file: main.bbl
\begin{thebibliography}{59}
\providecommand{\natexlab}[1]{#1}
\providecommand{\url}[1]{\texttt{#1}}
\expandafter\ifx\csname urlstyle\endcsname\relax
  \providecommand{\doi}[1]{doi: #1}\else
  \providecommand{\doi}{doi: \begingroup \urlstyle{rm}\Url}\fi

\bibitem[Avrahami et~al.(2023)Avrahami, Hayes, Gafni, Gupta, Taigman, Parikh, Lischinski, Fried, and Yin]{Avrahami_2023_CVPR_spatext}
Omri Avrahami, Thomas Hayes, Oran Gafni, Sonal Gupta, Yaniv Taigman, Devi Parikh, Dani Lischinski, Ohad Fried, and Xi Yin.
\newblock Spatext: Spatio-textual representation for controllable image generation.
\newblock In \emph{Proceedings of the IEEE/CVF Conference on Computer Vision and Pattern Recognition (CVPR)}, pages 18370--18380, 2023.

\bibitem[Brock et~al.(2019)Brock, Donahue, and Simonyan]{BigGAN}
Andrew Brock, Jeff Donahue, and Karen Simonyan.
\newblock Large scale {GAN} training for high fidelity natural image synthesis.
\newblock In \emph{7th International Conference on Learning Representations, {ICLR} 2019, New Orleans, LA, USA, May 6-9, 2019}. OpenReview.net, 2019.

\bibitem[Chen et~al.(2020)Chen, Hu, Gavves, Mensink, Mettes, Yang, and Snoek]{PointMixup}
Yunlu Chen, Tao Hu, Efstratios Gavves, Thomas Mensink, Pascal Mettes, Pengwan Yang, and Cees Snoek.
\newblock \emph{PointMixup: Augmentation for Point Clouds}, pages 330--345.
\newblock 2020.

\bibitem[Dai et~al.(2017)Dai, Chang, Savva, Halber, Funkhouser, and Nie{\ss}ner]{dai2017scannet}
Angela Dai, Angel~X. Chang, Manolis Savva, Maciej Halber, Thomas Funkhouser, and Matthias Nie{\ss}ner.
\newblock Scannet: Richly-annotated 3d reconstructions of indoor scenes.
\newblock In \emph{Proc. Computer Vision and Pattern Recognition (CVPR), IEEE}, 2017.

\bibitem[Dong and Lin(2023)]{dong2023weakly}
Shichao Dong and Guosheng Lin.
\newblock Weakly supervised 3d instance segmentation without instance-level annotations, 2023.

\bibitem[Dong et~al.(2022)Dong, Lin, and Hung]{dong2022learning}
Shichao Dong, Guosheng Lin, and Tzu-Yi Hung.
\newblock Learning regional purity for instance segmentation on 3d point clouds.
\newblock In \emph{European Conference on Computer Vision}, pages 56--72. Springer, 2022.

\bibitem[Dong et~al.(2023{\natexlab{a}})Dong, Li, Wei, Liu, and Lin]{dong2023collaborative}
Shichao Dong, Ruibo Li, Jiacheng Wei, Fayao Liu, and Guosheng Lin.
\newblock Collaborative propagation on multiple instance graphs for 3d instance segmentation with single-point supervision.
\newblock In \emph{Proceedings of the IEEE/CVF International Conference on Computer Vision}, pages 16665--16674, 2023{\natexlab{a}}.

\bibitem[Dong et~al.(2023{\natexlab{b}})Dong, Liu, and Lin]{dong2023leveraging}
Shichao Dong, Fayao Liu, and Guosheng Lin.
\newblock Leveraging large-scale pretrained vision foundation models for label-efficient 3d point cloud segmentation.
\newblock \emph{arXiv preprint arXiv:2311.01989}, 2023{\natexlab{b}}.

\bibitem[Gafni et~al.(2022)Gafni, Polyak, Ashual, Sheynin, Parikh, and Taigman]{make_a_scene}
Oran Gafni, Adam Polyak, Oron Ashual, Shelly Sheynin, Devi Parikh, and Yaniv Taigman.
\newblock Make-a-scene: Scene-based text-to-image generation with human priors, 2022.

\bibitem[Goodfellow et~al.(2014)Goodfellow, Pouget-Abadie, Mirza, Xu, Warde-Farley, Ozair, Courville, and Bengio]{GAN}
Ian Goodfellow, Jean Pouget-Abadie, Mehdi Mirza, Bing Xu, David Warde-Farley, Sherjil Ozair, Aaron Courville, and Yoshua Bengio.
\newblock Generative adversarial nets.
\newblock In \emph{Advances in Neural Information Processing Systems}. Curran Associates, Inc., 2014.

\bibitem[Graham and van~der Maaten(2017)]{Submanifold}
Benjamin Graham and Laurens van~der Maaten.
\newblock Submanifold sparse convolutional networks.
\newblock \emph{CoRR}, abs/1706.01307, 2017.

\bibitem[Graham et~al.(2018)Graham, Engelcke, and van~der Maaten]{3DSemanticSegmentationWithSubmanifoldSparseConvNet}
Benjamin Graham, Martin Engelcke, and Laurens van~der Maaten.
\newblock 3d semantic segmentation with submanifold sparse convolutional networks.
\newblock \emph{CVPR}, 2018.

\bibitem[He and Wang(2023)]{openlrm}
Zexin He and Tengfei Wang.
\newblock Openlrm: Open-source large reconstruction models.
\newblock \url{https://github.com/3DTopia/OpenLRM}, 2023.

\bibitem[Ho et~al.(2020)Ho, Jain, and Abbeel]{DDPM}
Jonathan Ho, Ajay Jain, and Pieter Abbeel.
\newblock Denoising diffusion probabilistic models.
\newblock In \emph{Advances in Neural Information Processing Systems 33: Annual Conference on Neural Information Processing Systems 2020, NeurIPS 2020, December 6-12, 2020, virtual}, 2020.

\bibitem[Hong et~al.(2023)Hong, Zhang, Gu, Bi, Zhou, Liu, Liu, Sunkavalli, Bui, and Tan]{hong2023lrm}
Yicong Hong, Kai Zhang, Jiuxiang Gu, Sai Bi, Yang Zhou, Difan Liu, Feng Liu, Kalyan Sunkavalli, Trung Bui, and Hao Tan.
\newblock Lrm: Large reconstruction model for single image to 3d.
\newblock \emph{arXiv preprint arXiv:2311.04400}, 2023.

\bibitem[Hu et~al.(2022)Hu, Shen, Wallis, Allen-Zhu, Li, Wang, Wang, and Chen]{hu2022lora}
Edward~J Hu, Yelong Shen, Phillip Wallis, Zeyuan Allen-Zhu, Yuanzhi Li, Shean Wang, Lu Wang, and Weizhu Chen.
\newblock Lo{RA}: Low-rank adaptation of large language models.
\newblock In \emph{International Conference on Learning Representations}, 2022.

\bibitem[Huang et~al.(2024)Huang, Huang, Liu, Yan, Lv, Liu, Xiong, Zhang, Chen, and Cao]{huang2024diffusion}
Yi Huang, Jiancheng Huang, Yifan Liu, Mingfu Yan, Jiaxi Lv, Jianzhuang Liu, Wei Xiong, He Zhang, Shifeng Chen, and Liangliang Cao.
\newblock Diffusion model-based image editing: A survey.
\newblock \emph{arXiv preprint arXiv:2402.17525}, 2024.

\bibitem[Jiang et~al.(2020)Jiang, Zhao, Shi, Liu, Fu, and Jia]{pointgroup}
Li Jiang, Hengshuang Zhao, Shaoshuai Shi, Shu Liu, Chi-Wing Fu, and Jiaya Jia.
\newblock Pointgroup: Dual-set point grouping for 3d instance segmentation, 2020.

\bibitem[Kerbl et~al.(2023)Kerbl, Kopanas, Leimk{\"u}hler, and Drettakis]{kerbl3Dgaussians}
Bernhard Kerbl, Georgios Kopanas, Thomas Leimk{\"u}hler, and George Drettakis.
\newblock 3d gaussian splatting for real-time radiance field rendering.
\newblock \emph{ACM Transactions on Graphics}, 42\penalty0 (4), 2023.

\bibitem[Kim et~al.(2021)Kim, Lee, Hwang, Lee, Hwang, and Kim]{Kim_2021_ICCV}
Sihyeon Kim, Sanghyeok Lee, Dasol Hwang, Jaewon Lee, Seong~Jae Hwang, and Hyunwoo~J. Kim.
\newblock Point cloud augmentation with weighted local transformations.
\newblock In \emph{Proceedings of the IEEE/CVF International Conference on Computer Vision (ICCV)}, pages 548--557, 2021.

\bibitem[Kingma and Welling(2013)]{VAE}
Diederik~P Kingma and Max Welling.
\newblock Auto-encoding variational bayes.
\newblock \emph{arXiv preprint arXiv:1312.6114}, 2013.

\bibitem[Lee et~al.(2021)Lee, Lee, Lee, Lee, Lee, Woo, and Lee]{lee2021regularization}
Dogyoon Lee, Jaeha Lee, Junhyeop Lee, Hyeongmin Lee, Minhyeok Lee, Sungmin Woo, and Sangyoun Lee.
\newblock Regularization strategy for point cloud via rigidly mixed sample.
\newblock In \emph{Proceedings of the IEEE/CVF Conference on Computer Vision and Pattern Recognition}, pages 15900--15909, 2021.

\bibitem[Lee et~al.(2022)Lee, Jeon, Kim, Xiong, and Kim]{lee2022sagemix}
Sanghyeok Lee, Minkyu Jeon, Injae Kim, Yunyang Xiong, and Hyunwoo~J. Kim.
\newblock Sagemix: Saliency-guided mixup for point clouds.
\newblock In \emph{Advances in Neural Information Processing Systems}, 2022.

\bibitem[Li et~al.(2019)Li, Li, Fu, Cohen-Or, and Heng]{li2019pugan}
Ruihui Li, Xianzhi Li, Chi-Wing Fu, Daniel Cohen-Or, and Pheng-Ann Heng.
\newblock Pu-gan: a point cloud upsampling adversarial network.
\newblock In \emph{{IEEE} International Conference on Computer Vision ({ICCV})}, 2019.

\bibitem[Li et~al.(2020)Li, Li, Heng, and Fu]{li2020pointaugment}
Ruihui Li, Xianzhi Li, Pheng-Ann Heng, and Chi-Wing Fu.
\newblock {PointAugment}: An auto-augmentation framework for point cloud classification.
\newblock In \emph{Proceedings of the IEEE/CVF Conference on Computer Vision and Pattern Recognition (CVPR)}, pages 6378--6387, 2020.

\bibitem[Li et~al.(2023)Li, Liu, Wu, Mu, Yang, Gao, Li, and Lee]{li2023gligen}
Yuheng Li, Haotian Liu, Qingyang Wu, Fangzhou Mu, Jianwei Yang, Jianfeng Gao, Chunyuan Li, and Yong~Jae Lee.
\newblock Gligen: Open-set grounded text-to-image generation.
\newblock \emph{CVPR}, 2023.

\bibitem[Liu et~al.(2023{\natexlab{a}})Liu, Shi, Chen, Zhang, Xu, Wei, Chen, Zeng, Gu, and Su]{liu2023one2345++}
Minghua Liu, Ruoxi Shi, Linghao Chen, Zhuoyang Zhang, Chao Xu, Xinyue Wei, Hansheng Chen, Chong Zeng, Jiayuan Gu, and Hao Su.
\newblock One-2-3-45++: Fast single image to 3d objects with consistent multi-view generation and 3d diffusion.
\newblock \emph{arXiv preprint arXiv:2311.07885}, 2023{\natexlab{a}}.

\bibitem[Liu et~al.(2023{\natexlab{b}})Liu, Xu, Jin, Chen, Xu, Su, et~al.]{liu2023one2345}
Minghua Liu, Chao Xu, Haian Jin, Linghao Chen, Zexiang Xu, Hao Su, et~al.
\newblock One-2-3-45: Any single image to 3d mesh in 45 seconds without per-shape optimization.
\newblock \emph{arXiv preprint arXiv:2306.16928}, 2023{\natexlab{b}}.

\bibitem[Liu et~al.(2023{\natexlab{c}})Liu, Wu, Hoorick, Tokmakov, Zakharov, and Vondrick]{liu2023zero1to3}
Ruoshi Liu, Rundi Wu, Basile~Van Hoorick, Pavel Tokmakov, Sergey Zakharov, and Carl Vondrick.
\newblock Zero-1-to-3: Zero-shot one image to 3d object, 2023{\natexlab{c}}.

\bibitem[Long et~al.(2023)Long, Guo, Lin, Liu, Dou, Liu, Ma, Zhang, Habermann, Theobalt, et~al.]{long2023wonder3d}
Xiaoxiao Long, Yuan-Chen Guo, Cheng Lin, Yuan Liu, Zhiyang Dou, Lingjie Liu, Yuexin Ma, Song-Hai Zhang, Marc Habermann, Christian Theobalt, et~al.
\newblock Wonder3d: Single image to 3d using cross-domain diffusion.
\newblock \emph{arXiv preprint arXiv:2310.15008}, 2023.

\bibitem[Melas-Kyriazi et~al.(2023)Melas-Kyriazi, Rupprecht, Laina, and Vedaldi]{melaskyriazi2023realfusion}
Luke Melas-Kyriazi, Christian Rupprecht, Iro Laina, and Andrea Vedaldi.
\newblock Realfusion: 360 reconstruction of any object from a single image.
\newblock In \emph{CVPR}, 2023.

\bibitem[Mildenhall et~al.(2020)Mildenhall, Srinivasan, Tancik, Barron, Ramamoorthi, and Ng]{mildenhall2020nerf}
Ben Mildenhall, Pratul~P. Srinivasan, Matthew Tancik, Jonathan~T. Barron, Ravi Ramamoorthi, and Ren Ng.
\newblock Nerf: Representing scenes as neural radiance fields for view synthesis.
\newblock In \emph{ECCV}, 2020.

\bibitem[Mou et~al.(2023)Mou, Wang, Xie, Wu, Zhang, Qi, Shan, and Qie]{mou2023t2i}
Chong Mou, Xintao Wang, Liangbin Xie, Yanze Wu, Jian Zhang, Zhongang Qi, Ying Shan, and Xiaohu Qie.
\newblock T2i-adapter: Learning adapters to dig out more controllable ability for text-to-image diffusion models.
\newblock \emph{arXiv preprint arXiv:2302.08453}, 2023.

\bibitem[Nekrasov et~al.(2021)Nekrasov, Schult, Litany, Leibe, and Engelmann]{Mix3D_Nekrasov213DV}
Alexey Nekrasov, Jonas Schult, Or Litany, Bastian Leibe, and Francis Engelmann.
\newblock {Mix3D: Out-of-Context Data Augmentation for 3D Scenes}.
\newblock In \emph{{International Conference on 3D Vision (3DV)}}, 2021.

\bibitem[Nichol and Dhariwal(2021)]{ImprovedDDPM}
Alexander~Quinn Nichol and Prafulla Dhariwal.
\newblock Improved denoising diffusion probabilistic models.
\newblock In \emph{Proceedings of the 38th International Conference on Machine Learning, {ICML} 2021, 18-24 July 2021, Virtual Event}, pages 8162--8171. {PMLR}, 2021.

\bibitem[Nichol et~al.(2022)Nichol, Dhariwal, Ramesh, Shyam, Mishkin, McGrew, Sutskever, and Chen]{GLIDE}
Alexander~Quinn Nichol, Prafulla Dhariwal, Aditya Ramesh, Pranav Shyam, Pamela Mishkin, Bob McGrew, Ilya Sutskever, and Mark Chen.
\newblock {GLIDE:} towards photorealistic image generation and editing with text-guided diffusion models.
\newblock In \emph{International Conference on Machine Learning, {ICML} 2022, 17-23 July 2022, Baltimore, Maryland, {USA}}, pages 16784--16804. {PMLR}, 2022.

\bibitem[Pan et~al.(2023)Pan, Tewari, Leimk{\"u}hler, Liu, Meka, and Theobalt]{pan2023draggan}
Xingang Pan, Ayush Tewari, Thomas Leimk{\"u}hler, Lingjie Liu, Abhimitra Meka, and Christian Theobalt.
\newblock Drag your gan: Interactive point-based manipulation on the generative image manifold.
\newblock In \emph{ACM SIGGRAPH 2023 Conference Proceedings}, 2023.

\bibitem[Poole et~al.(2022)Poole, Jain, Barron, and Mildenhall]{Poole2022DreamFusionTU}
Ben Poole, Ajay Jain, Jonathan~T. Barron, and Ben Mildenhall.
\newblock Dreamfusion: Text-to-3d using 2d diffusion.
\newblock \emph{ArXiv}, abs/2209.14988, 2022.

\bibitem[Qi et~al.(2019)Qi, Litany, He, and Guibas]{qi2019deep}
Charles~R Qi, Or Litany, Kaiming He, and Leonidas~J Guibas.
\newblock Deep hough voting for 3d object detection in point clouds.
\newblock In \emph{Proceedings of the IEEE International Conference on Computer Vision}, 2019.

\bibitem[Qian et~al.(2024)Qian, Mai, Hamdi, Ren, Siarohin, Li, Lee, Skorokhodov, Wonka, Tulyakov, and Ghanem]{Magic123}
Guocheng Qian, Jinjie Mai, Abdullah Hamdi, Jian Ren, Aliaksandr Siarohin, Bing Li, Hsin-Ying Lee, Ivan Skorokhodov, Peter Wonka, Sergey Tulyakov, and Bernard Ghanem.
\newblock Magic123: One image to high-quality 3d object generation using both 2d and 3d diffusion priors.
\newblock In \emph{The Twelfth International Conference on Learning Representations (ICLR)}, 2024.

\bibitem[Ramesh et~al.(2022)Ramesh, Dhariwal, Nichol, Chu, and Chen]{unCLIP}
Aditya Ramesh, Prafulla Dhariwal, Alex Nichol, Casey Chu, and Mark Chen.
\newblock Hierarchical text-conditional image generation with clip latents, 2022.

\bibitem[Razavi et~al.(2019)Razavi, van~den Oord, and Vinyals]{VQ-VAE-2}
Ali Razavi, A{\"{a}}ron van~den Oord, and Oriol Vinyals.
\newblock Generating diverse high-fidelity images with {VQ-VAE-2}.
\newblock In \emph{Advances in Neural Information Processing Systems 32: Annual Conference on Neural Information Processing Systems 2019, NeurIPS 2019, December 8-14, 2019, Vancouver, BC, Canada}, pages 14837--14847, 2019.

\bibitem[Rombach et~al.(2022)Rombach, Blattmann, Lorenz, Esser, and Ommer]{StableDiffusion}
Robin Rombach, Andreas Blattmann, Dominik Lorenz, Patrick Esser, and Bj{\"{o}}rn Ommer.
\newblock High-resolution image synthesis with latent diffusion models.
\newblock In \emph{{IEEE/CVF} Conference on Computer Vision and Pattern Recognition, {CVPR} 2022, New Orleans, LA, USA, June 18-24, 2022}, pages 10674--10685. {IEEE}, 2022.

\bibitem[Saharia et~al.(2022)Saharia, Chan, Saxena, Li, Whang, Denton, Ghasemipour, Ayan, Mahdavi, Lopes, Salimans, Ho, Fleet, and Norouzi]{Imagen}
Chitwan Saharia, William Chan, Saurabh Saxena, Lala Li, Jay Whang, Emily Denton, Seyed Kamyar~Seyed Ghasemipour, Burcu~Karagol Ayan, S.~Sara Mahdavi, Rapha~Gontijo Lopes, Tim Salimans, Jonathan Ho, David~J Fleet, and Mohammad Norouzi.
\newblock Photorealistic text-to-image diffusion models with deep language understanding, 2022.

\bibitem[Shen et~al.(2023)Shen, Yang, and Wang]{shen2023anything3d}
Qiuhong Shen, Xingyi Yang, and Xinchao Wang.
\newblock Anything-3d: Towards single-view anything reconstruction in the wild, 2023.

\bibitem[Shi et~al.(2023{\natexlab{a}})Shi, Chen, Zhang, Liu, Xu, Wei, Chen, Zeng, and Su]{shi2023zero123plus}
Ruoxi Shi, Hansheng Chen, Zhuoyang Zhang, Minghua Liu, Chao Xu, Xinyue Wei, Linghao Chen, Chong Zeng, and Hao Su.
\newblock Zero123++: a single image to consistent multi-view diffusion base model, 2023{\natexlab{a}}.

\bibitem[Shi et~al.(2019)Shi, Wang, and Li]{Shi_2019_CVPR}
Shaoshuai Shi, Xiaogang Wang, and Hongsheng Li.
\newblock Pointrcnn: 3d object proposal generation and detection from point cloud.
\newblock In \emph{The IEEE Conference on Computer Vision and Pattern Recognition (CVPR)}, 2019.

\bibitem[Shi et~al.(2023{\natexlab{b}})Shi, Xue, Pan, Zhang, Tan, and Bai]{shi2023dragdiffusion}
Yujun Shi, Chuhui Xue, Jiachun Pan, Wenqing Zhang, Vincent~YF Tan, and Song Bai.
\newblock Dragdiffusion: Harnessing diffusion models for interactive point-based image editing.
\newblock \emph{arXiv preprint arXiv:2306.14435}, 2023{\natexlab{b}}.

\bibitem[Sohl{-}Dickstein et~al.(2015)Sohl{-}Dickstein, Weiss, Maheswaranathan, and Ganguli]{OriginalDiffusionPaper}
Jascha Sohl{-}Dickstein, Eric~A. Weiss, Niru Maheswaranathan, and Surya Ganguli.
\newblock Deep unsupervised learning using nonequilibrium thermodynamics.
\newblock In \emph{Proceedings of the 32nd International Conference on Machine Learning, {ICML} 2015, Lille, France, 6-11 July 2015}, pages 2256--2265. JMLR.org, 2015.

\bibitem[Song et~al.(2021)Song, Meng, and Ermon]{DDIM}
Jiaming Song, Chenlin Meng, and Stefano Ermon.
\newblock Denoising diffusion implicit models.
\newblock In \emph{9th International Conference on Learning Representations, {ICLR} 2021, Virtual Event, Austria, May 3-7, 2021}. OpenReview.net, 2021.

\bibitem[Tang et~al.(2023)Tang, Wang, Zhang, Zhang, Yi, Ma, and Chen]{Tang_2023_ICCV}
Junshu Tang, Tengfei Wang, Bo Zhang, Ting Zhang, Ran Yi, Lizhuang Ma, and Dong Chen.
\newblock Make-it-3d: High-fidelity 3d creation from a single image with diffusion prior.
\newblock In \emph{Proceedings of the IEEE/CVF International Conference on Computer Vision (ICCV)}, pages 22819--22829, 2023.

\bibitem[Tang et~al.(2024)Tang, Chen, Chen, Wang, Zeng, and Liu]{tang2024lgm}
Jiaxiang Tang, Zhaoxi Chen, Xiaokang Chen, Tengfei Wang, Gang Zeng, and Ziwei Liu.
\newblock Lgm: Large multi-view gaussian model for high-resolution 3d content creation.
\newblock \emph{arXiv preprint arXiv:2402.05054}, 2024.

\bibitem[Vu et~al.(2023)Vu, Sarkar, Zhang, Hua, and Yeung]{vu2023testtime}
Tuan-Anh Vu, Srinjay Sarkar, Zhiyuan Zhang, Binh-Son Hua, and Sai-Kit Yeung.
\newblock Test-time augmentation for 3d point cloud classification and segmentation, 2023.

\bibitem[Wang et~al.(2021)Wang, Liu, Liu, Theobalt, Komura, and Wang]{wang2021neus}
Peng Wang, Lingjie Liu, Yuan Liu, Christian Theobalt, Taku Komura, and Wenping Wang.
\newblock Neus: Learning neural implicit surfaces by volume rendering for multi-view reconstruction.
\newblock \emph{arXiv preprint arXiv:2106.10689}, 2021.

\bibitem[Yang et~al.(2024)Yang, Kang, Huang, Xu, Feng, and Zhao]{depthanything}
Lihe Yang, Bingyi Kang, Zilong Huang, Xiaogang Xu, Jiashi Feng, and Hengshuang Zhao.
\newblock Depth anything: Unleashing the power of large-scale unlabeled data.
\newblock In \emph{CVPR}, 2024.

\bibitem[Zhang et~al.(2023)Zhang, Rao, and Agrawala]{zhang2023adding_controlnet}
Lvmin Zhang, Anyi Rao, and Maneesh Agrawala.
\newblock Adding conditional control to text-to-image diffusion models, 2023.

\bibitem[Zhang et~al.(2021)Zhang, Chen, Ling, Gao, Zhang, Torralba, and Fidler]{StyleGANRender}
Yuxuan Zhang, Wenzheng Chen, Huan Ling, Jun Gao, Yinan Zhang, Antonio Torralba, and Sanja Fidler.
\newblock Image gans meet differentiable rendering for inverse graphics and interpretable 3d neural rendering.
\newblock In \emph{9th International Conference on Learning Representations, {ICLR} 2021, Virtual Event, Austria, May 3-7, 2021}. OpenReview.net, 2021.

\bibitem[Zhu et~al.(2024)Zhu, Fan, and Weng]{zhu2024advancements}
Qinfeng Zhu, Lei Fan, and Ningxin Weng.
\newblock Advancements in point cloud data augmentation for deep learning: A survey, 2024.

\bibitem[Zou et~al.(2023)Zou, Yu, Guo, Li, Liang, Cao, and Zhang]{TriplanMeetsGaussian}
Zi-Xin Zou, Zhipeng Yu, Yuan-Chen Guo, Yangguang Li, Ding Liang, Yan-Pei Cao, and Song-Hai Zhang.
\newblock Triplane meets gaussian splatting: Fast and generalizable single-view 3d reconstruction with transformers.
\newblock \emph{arXiv preprint arXiv:2312.09147}, 2023.

\end{thebibliography}


\begin{thebibliography}{3}
\providecommand{\natexlab}[1]{#1}
\providecommand{\url}[1]{\texttt{#1}}
\expandafter\ifx\csname urlstyle\endcsname\relax
  \providecommand{\doi}[1]{doi: #1}\else
  \providecommand{\doi}{doi: \begingroup \urlstyle{rm}\Url}\fi

\bibitem[Dai et~al.(2017)Dai, Chang, Savva, Halber, Funkhouser, and Nie{\ss}ner]{dai2017scannet}
Angela Dai, Angel~X. Chang, Manolis Savva, Maciej Halber, Thomas Funkhouser, and Matthias Nie{\ss}ner.
\newblock Scannet: Richly-annotated 3d reconstructions of indoor scenes.
\newblock In \emph{Proc. Computer Vision and Pattern Recognition (CVPR), IEEE}, 2017.

\bibitem[Graham and van~der Maaten(2017)]{Submanifold}
Benjamin Graham and Laurens van~der Maaten.
\newblock Submanifold sparse convolutional networks.
\newblock \emph{CoRR}, abs/1706.01307, 2017.

\bibitem[Jiang et~al.(2020)Jiang, Zhao, Shi, Liu, Fu, and Jia]{pointgroup}
Li Jiang, Hengshuang Zhao, Shaoshuai Shi, Shu Liu, Chi-Wing Fu, and Jiaya Jia.
\newblock Pointgroup: Dual-set point grouping for 3d instance segmentation, 2020.

\end{thebibliography}
